%% file: paper_main.tex
\pdfoutput=1

\documentclass[11pt]{article}

\usepackage[final]{acl}
\usepackage{times}
\usepackage{latexsym}

\usepackage[T1]{fontenc}

\usepackage[utf8]{inputenc}

\usepackage{microtype}

\usepackage{inconsolata}

\usepackage{graphicx}
\usepackage{microtype}
\usepackage{hyperref}
\usepackage{url}
\usepackage{booktabs}
\usepackage{enumitem}
\usepackage{multirow}
\usepackage{float}
\usepackage{xspace}
\usepackage{makecell}
\usepackage{wrapfig}
\usepackage{tabularx}
\usepackage{subcaption}
\usepackage{setspace}
\usepackage{color} 
\usepackage{longtable}
\usepackage{tcolorbox}

\definecolor{darkblue}{rgb}{0, 0, 0.5}
\definecolor{darkgreen}{RGB}{0,100,0}
\hypersetup{colorlinks,linkcolor={darkblue},citecolor={darkblue},urlcolor={darkblue}}
\title{A Practical Examination of AI-Generated Text Detectors for Large Language Models}

\author{Brian Tufts\\
Carnegie Mellon University\\
\texttt{btufts@cs.cmu.edu}\\
\And
Xuandong Zhao\\
UC Berkeley\\
\texttt{xuandongzhao@berkeley.edu}\\
\And
Lei Li\\
Carnegie Mellon University\\
\texttt{leili@cs.cmu.edu}
}

\begin{document}
\maketitle

\begin{abstract}
\input{sections/00_abs}

\end{abstract}

\section{Introduction}
\label{sec:intro}
\input{sections/01_intro}

\section{Related Work and Background} \label{section:related-work}

\input{sections/02_related}

\section{Benchmarking Procedure}
\label{sec:method}

\input{sections/03_method}

\section{Experiment}
\label{sec:exp}

\input{sections/04_experiment}

\section{Results and Analysis}
\label{sec:result}

\input{sections/05_results}

\section{Conclusion}
\label{sec:conclusion}
\input{sections/07_conclusion}

\section{Limitations}
\label{sec:limit}
\input{sections/09_limitations}

\section*{Acknowledgements} L.L. is partly supported by a gift from Hydrox AI and a CMU CyLab Seed Fund. The views expressed are those of the author and do not reflect the official policy or position of the funding agencies.


\bibliography{clean}
\newpage
\appendix
~
\newpage

\section{More Results}
\label{sec:more}
\input{sections/08_appendix}

\end{document}

%% file: sections/00_abs.tex
The proliferation of large language models has raised growing concerns about their misuse, particularly in cases where AI-generated text is falsely attributed to human authors. Machine-generated content detectors claim to effectively identify such text under various conditions and from any language model. This paper critically evaluates these claims by assessing several popular detectors (RADAR, Wild, T5Sentinel, Fast-DetectGPT, PHD, LogRank, Binoculars) on a range of domains, datasets, and models that these detectors have not previously encountered. We employ various prompting strategies to simulate practical adversarial attacks, demonstrating that even moderate efforts can significantly evade detection. We emphasize the importance of the true positive rate at a specific false positive rate (TPR@FPR) metric and demonstrate that these detectors perform poorly in certain settings, with TPR@.01 as low as 0\%. Our findings suggest that both trained and zero-shot detectors struggle to maintain high sensitivity while achieving a reasonable true positive rate. All code and data necessary to reproduce our experiments are available at \url{https://github.com/LeiLiLab/llm-detector-eval}.

%% file: sections/01_intro.tex


Large language models (LLMs) are becoming increasingly accessible and powerful, leading to numerous beneficial applications \citep{touvron2023llama, achiam2023gpt}. However, they also pose risks if used maliciously, such as generating fake news articles, facilitating academic plagiarism or spam content \citep{feng2024does, zellers2019defending, perkins2023academic, fraser2024detectingaigeneratedtextfactors}. The potential for misuse of LLMs has become a significant concern for major tech corporations, particularly in light of the 2024 elections in the united states. At the Munich Security Conference on February 16th, 2024, these companies pledged to combat misleading machine-generated content, acknowledging the potential of AI to deceptively influence electoral outcomes \citep{aielectionsaccord}. As a result, there is a growing need to develop reliable methods for differentiating between LLM-generated and human-written content. To ensure the effectiveness and accountability of LLM detection methods, continuous evaluation of popular techniques is crucial.

Many methods have been released recently that claim to have a strong ability to detect the difference between AI-generated and human-generated texts. These detectors primarily fall into three categories: trained detectors, zero-shot detectors, and watermarking techniques \citep{yang2023survey, ghosal2023towards, tang2023science}. \emph{Trained detectors} utilize datasets of human and AI-generated texts and train a binary classification model to detect the source of a text \citep{zellers2019defending, hovy2016enemy, hu2023radar, tian2023gptzero, verma2023ghostbuster}. \emph{Zero-shot detection} utilizes a language model's inherent traits to identify text it generates, without explicit training for detection tasks other than calibrating a threshold for detection in some cases \citep{ gehrmann2019gltr, mitchell2023detectgpt, bao2024fast, yang2023dna,venkatraman2023gpt}. \emph{Watermarking} is another technique in which the model owner embeds a specific probabilistic pattern into the text to make it detectable \citet{kirchenbauer2023watermark}. However, watermarking requires the model owner to add the signal, and its design has theoretical guarantees; we do not evaluate watermarking models in this study.

In this paper, we test the robustness of these detection methods to unseen models, data sources, and adversarial prompting. To do this, we treat all model-generated text as a black box generation. That is, none of the detectors know the source of the text or have access to the model generating the text. This presents the most realistic scenario where the user is presented with text and wants to know if it is AI-generated or not. Our contributions can be summarized as follows:

\begin{itemize}[leftmargin=*]
\item We conduct a thorough evaluation of AI-generated text detectors on unseen models and tasks, providing insights into their effectiveness in real-world settings.
\item We analyze the performance of various detectors under practical adversarial prompting, exploring the extent to which prompting can be used to evade detection.
\item We demonstrate that high AUROC scores, which are often used as a measure of performance in classification tasks, do not necessarily translate to practical usage for machine-generated text detection. Instead, we motivate using the metric of true positive rate (TPR) at a 1\% false positive rate (FPR) threshold as a more reliable indicator of a detector's effectiveness in practice.
\end{itemize}



%% file: sections/02_related.tex
There is a variety of related work that discusses text detectors. These works cover different aspects, such as the text detectors themselves, their types, evaluation, and red-teaming of detectors.

\paragraph{Text Detectors.} Machine-generated text detectors can be divided into trained classifiers, zero-shot classifiers, and watermark methods~\citep{yang2023survey, hans2024spotting, ghosal2023towards, jawahar2020automatic}. (1) Trained detectors use classification models to determine if the text is machine-generated or human-written~\citep{zellers2019defending, hovy2016enemy, hu2023radar, tian2023gptzero, verma2023ghostbuster}. However, the increasing prevalence of machine-generated content~\citep{europol} makes it difficult to label human-generated work for training, as even humans find it hard to distinguish between the two~\citep{darda2023value}. (2) Zero-shot detectors leverage intrinsic statistical differences between machine-generated and human-generated text~\citep{gehrmann2019gltr, mitchell2023detectgpt, bao2024fast, yang2023dna, venkatraman2023gpt}. Proposed methods include using entropy~\citep{lavergne2008detecting}, log probability~\citep{solaiman2019release}, and more recently, intrinsic dimensionality~\citep{tulchinskii2024intrinsic}. (3) Watermark-based detection, introduced by \citet{kirchenbauer2023watermark}, involves embedding a hidden but detectable pattern in the generated output. Various enhancements to this method have been suggested (e.g., \citet{zhao2023provable, lee2023wrote}). This paper focuses on the black-box setting, which closely resembles real-world detection scenarios. Watermarking is not tested due to its guaranteed detectability and low false positive rates (e.g.,~\citep{zhao2023provable}). The primary concern is detecting un-watermarked text, as it is the most commonly encountered and poses the greatest threat.

\paragraph{Evaluation of Text Detectors.} The most commonly utilized metric in evaluating detectors is the area under the receiver operating curve (AUROC)~\citep{mitchell2023detectgpt, sadasivan2023can}. Although it offers a reasonable estimate of detector performance, research by \citet{krishna2024paraphrasing, yang2023dna}, and our experimental results demonstrate that there can be a substantial difference in performance between two models with AUROC values nearing the maximum of 1.0. Consequently, the true positive rate at a fixed false positive rate (TPR@FPR) presents a more accurate representation of a detector's practical effectiveness. Both AUROC and true positive rate at a fixed false positive are important metrics for a complete evaluation of text detectors.

\paragraph{Redteaming Language Model Detectors.} AI text detectors are increasingly evaluated in red teaming scenarios, with recent contributions from \citet{zhu2023promptbench, chakraborty2023counter, kumarage2023reliable, shi2024red, wang2024stumbling}. \citet{shi2024red} identifies two main evasion techniques: word substitution and instructional prompts. Word substitution includes query-based methods, which iteratively select low detection score substitutions, and query-free methods, which use random substitutions. Instructional prompts, akin to jailbreaking, instruct the model to mimic a human-written sample. Query-based word substitution proved most effective, reducing the True Positive Rate (TPR) to less than 5\% at a 40\% False Positive Rate (FPR) against DetectGPT. \citet{wang2024stumbling} explore robustness testing of language model detectors with three editing attacks: typo insertion, homoglyph alteration, and format character editing. Typo insertion adds typos, homoglyph alteration replaces characters with similar shapes, and format character editing uses invisible text disruptions. Paraphrasing attacks, noted by \citet{krishna2024paraphrasing}, include synonym substitution (model-free and model-assisted), span perturbations (masking and refilling random spans), and paraphrasing at sentence and text levels.


\begin{table}[t]
\centering
\resizebox{0.99\linewidth}{!}{
\begin{tabular}{lp{14.5cm}}
\toprule
\textbf{Method} & \textbf{Datasets} \\
\midrule
RADAR & OpenWebText Corpus~\citep{gokaslan2019openwebtext}, Xsum~\citep{narayan2018don}, SQuAD~\citep{rajpurkar2016squad}, Reddit Writing Prompts~\citep{fan2018hierarchical}, and TOEFL~\citep{liang2023gpt}\\
\midrule
Wild & Reddit CMV sub-community comments~\citep{tan2016winning}, Yelp Reviews~\citep{zhang2015character}, Xsum~\citep{narayan2018don}, TLDR\_news\footnote{\url{https://huggingface.co/datasets/JulesBelveze/tldr\_news}}, ELI5 dataset~\citep{fan2019eli5}, Reddit Writing Prompts~\citep{fan2018hierarchical}, ROCStories Corpora~\citep{mostafazadeh2016corpus}, HellaSwag~\citep{zellers2019hellaswag}, SQuAD~\citep{rajpurkar2016squad}, and SciGen~\citep{moosavi2021learning} \\
\midrule
T5Sentinel & OpenWebText Corpus~\citep{gokaslan2019openwebtext}\\
\midrule
Fast-DetectGPT & Xsum~\citep{narayan2018don}, SQuAD~\citep{rajpurkar2016squad}, Reddit Writing Prompts~\citep{fan2018hierarchical}, WMT16 English and German~\citep{bojar2016findings}, PubMedQA~\citep{jin2019pubmedqa} \\
\midrule
PHD & Wiki40b~\citep{guo2020wiki}, Reddit Writing Prompts~\citep{fan2018hierarchical}, WikiM~\citep{krishna2024paraphrasing}, StackExchange~\citep{tulchinskii2024intrinsic} \\
\midrule
LogRank & Xsum~\citep{narayan2018don}, SQuAD~\citep{rajpurkar2016squad}, Reddit Writing Prompts~\citep{fan2018hierarchical}\\
\midrule
Binoculars & CCNews~\citep{Hamborg2017news-please}, PubMed~\citep{Sen_Namata_Bilgic_Getoor_Galligher_Eliassi-Rad_2008}, CNN~\citep{Hermann}, ORCA~\citep{OpenOrca}\\
\bottomrule
\end{tabular}
}
\caption{Datasets used for training and evaluation by each model. To avoid data leakage and cherry-picking, these datasets are excluded from the current study.}
\label{tab:datasets}
\end{table}

\raggedbottom
\paragraph{Evaluated Detectors and Datasets.} In our paper, we evaluate seven representative detectors: RADAR~\citep{hu2023radar}, Detection in the Wild (Wild)~\citep{li2024mage}, T5Sentinel~\citep{chen2023token}, Fast-DetectGPT~\citep{bao2024fast}, PHD~\citep{tulchinskii2024intrinsic}, LogRank~\citep{ippolito-etal-2020-automatic}\footnote{LogRank has been evaluated on many datasets, we report the ones from \citet{mitchell2023detectgpt}.}, and Binoculars~\citep{hans2024spotting}. RADAR, Wild, and T5Sentinel are trained detectors, while Fast-DetectGPT, PHD, LogRank, and Binoculars are zero-shot detectors. To ensure a fair comparison and assess the detectors' ability to generalize to new data, we carefully select datasets that have not been used in the training or evaluation of these detectors.
Table \ref{tab:datasets} presents an overview of the datasets and domains on which each detector has been evaluated. Several datasets, such as Xsum, SQuAD, and Reddit Writing Prompts, have been used in the evaluation or training of multiple detectors. Although these detectors achieve strong Area Under the Receiver Operating Characteristic (AUROC) scores on these datasets, they do not report the True Positive Rate at a set False Positive Rate (TPR@FPR), which is a crucial metric in real-world scenarios. To address this gap, we aim to evaluate all seven detectors on the same datasets using both AUROC and TPR at FPR metrics.

\paragraph{Comparison to Previous Works.}
There are some other papers that have explored similar work to ours, specifically \citet{wang2024stumbling} and \citet{dugan2024raidsharedbenchmarkrobust}. Our work differs from theirs in some important ways. We do not focus as much on the various methods of red-teaming the detectors in complicated ways. Rather, we explore some more natural methods that an average person might utilize in practice. We also explore in more depth the variability in detector capabilities across various tasks and languages with discussion on potential sources of that difference. And lastly, we utilize newer models, which gives insight into the adaptability of the detectors.

%% file: sections/03_method.tex
Our benchmarking method involves compiling datasets that have not been encountered by any of the detectors during their training or evaluation phases. This approach ensures that the datasets represent new, unseen data and prevents the possibility of data contamination. For zero-shot detectors, this methodology eliminates the risk of using cherry-picked datasets that may bias the evaluation. For trained detectors, this reduces the risk of data leakage and tests on out-of-domain data. Furthermore, we assess the model's performance across a diverse range of domains that the detectors may not have been previously evaluated against. This comprehensive evaluation strategy allows for a more robust assessment of the detectors' generalization capabilities. Additionally, we evaluate the detectors on a variety of language models that they have not encountered before. This enables us to examine the detectors' performance on unfamiliar language models, providing a more comprehensive understanding of their effectiveness and adaptability.

\subsection{Datasets}
We evaluate each of the detectors on seven different tasks with three of the tasks, question answering, summarization, and dialogue writing, including multilingual results. The datasets chosen for each domain are as follows:
\begin{itemize}[leftmargin=*, ]
\item \textbf{Question Answering:} The MFAQ dataset \citep{de-bruyn-etal-2021-mfaq} was used for this domain. It contains over one million question-answer pairs in various languages. We used the English, Spanish, French, and Chinese subsets.
\item \textbf{Summarization:} We used the MTG summarization dataset \citep{chen2021mtg} for this task. The complete multilingual dataset comprises roughly 200k summarizations. We utilized the English, Spanish, French, and Chinese subsets.
\item \textbf{Dialogue Writing:} For this task, we utilized the MSAMSum dataset, a translated version of the SAMSum dataset\citep{feng-etal-2022-msamsum, gliwa2019samsum}. This dataset consists of over 16k dialogues with summaries in six languages. We utilized English, Spanish, French, and Chinese for consistency with the other multilingual domains.
\item \textbf{Code:} We used the APPS dataset \citep{hendrycks2021measuring}, which contains 10k code questions and solutions. The subset used was randomly selected from all the data included in APPS.
\item \textbf{Abstract Writing:} For this task, we utilized the Arxiv section of the scientific papers dataset \citep{cohan2018discourse} to avoid potential bias, as some detectors have previously been exposed to PubMed data. Additionally, we only selected papers published in 2020 or earlier to remove potential LLM influence.
\item \textbf{Review Writing:} The PeerRead dataset was used for the review writing task \citep{kang18naacl}. PeerRead contains over 10k peer reviews written by experts corresponding to the paper that they were written for.
\item \textbf{Translation:} We used the Par3 dataset \citep{Par3_2022}, which provides paragraph level translations from public-domain foreign language novels. Each paragraph includes at least 2 human translations of which we selected only one to represent human translation.
\end{itemize}

\subsection{Large Language Models}

Our objective is to evaluate the detectors on models that they have not previously been trained or assessed on to gauge their generalization capabilities. We evaluated 4 different models across every task. The models we use are Llama-3-Instruct 8B \citep{llama3modelcard}, Mistral-Instruct-v0.3 \citep{jiang2023mistral}, Phi-3-Mini-Instruct 4k \citep{abdin2024phi}, and GPT-4o.

\subsection{Detection Models}
The detection models were chosen from the newest and highest performing detectors in their respective categories. Our goal was to represent both trained and zero-shot detectors. As previously mentioned, the trained detectors we are using are RADAR \citep{hu2023radar}, Detection in the Wild (Wild) \citep{li2024mage}, and T5Sentinel \citep{chen2023token}. The zero-shot detectors we are using are Fast-DetectGPT \citep{bao2024fast}, GPTID \citep{tulchinskii2024intrinsic}, LogRank \citep{ippolito-etal-2020-automatic}, and Binoculars \citep{hans2024spotting}. Each of the zero-shot detectors utilize a generating model as a part of their detection process. We utilize the same underlying models as reported by each respective zero-shot model's original publication listed in Table \ref{tab:zeroshot-models}. We also evaluate every zero-shot method using three of the other underlying models for a more accurate comparison. This is notably unfair to the Binoculars method, which uses two different underlying models: base and instruction tuned. We replace both with the same model for these experiments because not all models have both base and instruction tuned versions.

Notably, we did not include any watermark detectors. The primary reason for this is that the evaluation techniques we use over various models would not work with watermark detection. While watermark detection has shown strong performance \citep{kirchenbauer2023watermark}, they have a significant drawback in that they only work if a model applies a watermark. In this paper, we assume a scenario in which no watermark is applied or it is unknown whether a watermark is applied. Therefore, we must turn to other detection methods.

\begin{table}[t]
\centering
\resizebox{0.99\linewidth}{!}{
\LARGE
\begin{tabular}{lp{14cm}}
\toprule
\textbf{Method} & \textbf{Model} \\
\midrule
Fast-DetectGPT & GPT-Neo-2.7B \citep{gpt-neo}\\
\midrule
GPTID & Roberta-Base \cite{liu2019robertarobustlyoptimizedbert}\\
\midrule
LogRank & GPT2-Medium \citep{radford2019language}\\
\midrule
Binoculars & Falcon-7B, Falcon-7B-Instruct \citep{almazrouei2023falconseriesopenlanguage}\\
\bottomrule
\end{tabular}}
\caption{Underlying models utilized by each zero-shot detection method.}
\label{tab:zeroshot-models}
\end{table}


\subsection{Evaluation Metrics}


In this study, we evaluate machine-generated text detectors using AUROC and TPR at a fixed FPR. Our findings, consistent with prior research \citep{krishna2024paraphrasing, yang2023dna}, suggest that AUROC alone may not reflect a detector's practical effectiveness, as a high AUROC score can still correspond to significant false positive rates. This is critical since false positives, particularly in fields like academia and media, can have severe consequences. We argue that TPR at a given FPR should be the standard evaluation metric, as demonstrated by a detector achieving a 0.89 AUROC but less than 20\% TPR at a 1\% FPR on a task.

\subsection{Red Teaming}
We employ two different methods of prompting for every task: plain prompting and adversarial prompting. Plain prompting involves using a typical assistant system prompt and providing the model with the same input that was given to the human for human-generated content. Adversarial prompting, on the other hand, requests that the model try to act more like a person. Examples of the question answering plain and adversarial prompts\footnote{The others can be found in the appendix Table \ref{tab:plain_prompts}.} are shown as follows:
\begin{tcolorbox}
[colback=gray!5!white,colframe=gray!75!black,title=\small Plain Prompt Example: Question Answering]
\small
You are a helpful question answering assistant that will answer a single question as completely as possible given the information in the question. Do NOT use any markdown, bullet, or numbered list formatting. The assistant will use ONLY paragraph formatting. **Respond only in \{language\}**.
\label{tab:plain_example}
\end{tcolorbox}

\begin{tcolorbox}[colback=gray!5!white,colframe=gray!75!black,title=\small Adversarial Prompt Example: Question Answering]
\small
\{Question answering prompt\} Try to sound as human as possible. 
\label{tab:template_example}
\end{tcolorbox}

We also conducted experiments using the LLMs as writing assistants. Specifically, we requested that the model rewrite the human response and improve upon its clarity and professionalism. This represents a scenario where a person will write down an answer first and then request that a model make their answer better before presenting it. The specific prompt we used it as follow:
\begin{tcolorbox}[colback=gray!5!white,colframe=gray!75!black,title=\small Rewriting Prompt]
\small
You are a helpful writing assistant. Rewrite the following text to improve clarity and professionalism. Do not provide any other text. Only provide the rewritten text.
\label{tab:rewrite_example}
\end{tcolorbox}


%% file: sections/04_experiment.tex
\raggedbottom
\subsection{Dataset Processing} Each dataset undergoes additional processing to prepare it for detection tasks. Research indicates that detectors of machine-generated text are more effective with longer content \citep{yang2023survey}. To leverage this, we aimed to use human samples of maximum possible length. However, the minimum length needed to obtain sufficient samples varied by task. We randomly selected 500 samples of human text from filtered subsets with the following token lengths using Llama2-13B tokenizer \citep{touvron2023llama}: 500 tokens for question answering, 400 tokens for code\footnote{\label{length_limit}Length limited to 2500 tokens.}, 150 tokens for summarization, 275 tokens for dialogue, 500 tokens for reviews, 500 tokens for abstracts, and 500 tokens for translation (Table \ref{tab:response_lenghts}). These 500 samples served as human examples. From them, prompts from the first 100 samples were chosen for use in the generator model, using the input given to the human author as the model prompt. This resulted in a dataset of 500 human examples and 100 machine-generated examples per model for a total of 400 machine-generated examples for each task. This slight data imbalance is intentional to ensure a more accurate TPR@FPR metric because there would likely be more human examples than machine generated examples in practice.

Detection methods show improved performance with longer text sequences \citep{wu2023survey} so we show the statistics of the text in Table \ref{tab:response_lenghts}. Our primary focus was on detectors' ability to identify AI-generated text while maintaining a low FPR. The longer length of human-generated text is likely to enhance the TPR@FPR by making it easier to detect as human. We considered the AI-generated text sufficiently long for two reasons. First, \citet{li2024mage} reports an average AI generation length of 279.99, which is much lower than our average token lengths. Their extensive training and evaluation data support the adequacy of this length for AI content. Second, our models, with a maximum generation length of 512 tokens \footnote{The averages can exceed this number due to different tokenizers and additional tokens to keep text coherent}, produced responses indicative of real-world lengths.


\begin{table}[t]
\small
\centering
\begin{tabular}{lcccc}
\toprule
\multirow{2}{*}{\textbf{Task}} & \multicolumn{2}{c}{\textbf{AI}} & \multicolumn{2}{c}{\textbf{Human}} \\ 
 & \textbf{Avg} & \textbf{Min} & \textbf{Avg} & \textbf{Min} \\ \midrule
Code & 486.58 & 15 & 4496.88 & 605 \\ 
QA & 508.01 & 24 & 1052.37 & 501 \\ 
Summ & 410.03 & 18 & 191.00 & 151 \\ 
Dialogue & 380.92 & 15 & 402.13 & 276 \\ 
Reviews & 551.28 & 24 & 796.06 & 501 \\ 
Abstract & 427.92 & 30 & 2081.88 & 501 \\ 
Translation & 525.32 & 256 & 772.75 & 501 \\ \bottomrule
\end{tabular}
\caption{Average and minimum token counts of machine-generated and human-generated text for each task, tokenized using the Llama2 tokenizer \citep{touvron2023llama}. Minimum token counts for human-generated text are omitted as they were previously described.}
\label{tab:response_lenghts}
\end{table}

\subsection{Text Generation and Detection Process} 
\begin{figure}[t]
    \centering
\includegraphics[width=.99\linewidth]{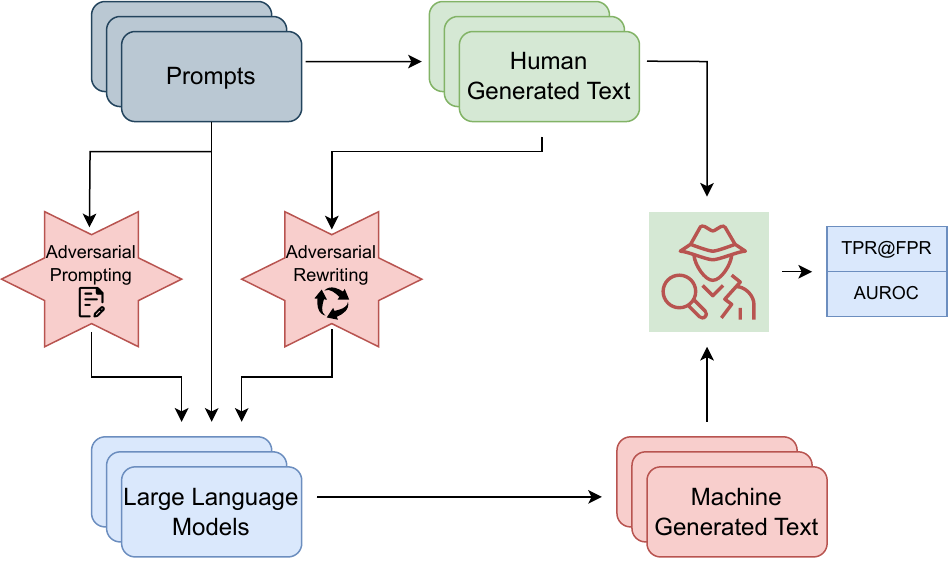} 
    \caption{Pipeline for prompting and evaluation. Adversarial prompting and rewriting are applied to the LLMs. After collecting machine-generated text, AUROC and TPR@FPR are measured for each detector.}
    \label{fig:png}
\end{figure}

Once the prompt samples were selected, we needed to generate positive examples. The process for this can be seen in Figure \ref{fig:png}. We employ three different strategies for prompting the models, one is a plain prompt and the other two are adversarial prompts. The first strategy involves using a basic prompt for each domain that explains the goal of the model and the desired output format. The second strategy consists of requesting that the model be as human as possible. The third strategy requests that the model rewrite and improve upon the human written response \footnote{Prompts and templates can be found in the appendix.}. The first strategy aims to simulate a basic system prompt that would generally be in place on a model someone is using to generate content. The second strategy simulates the case where a user might try to get the model to generate content that closely resembles human-generated content. The third strategy simulates a scenario where the user writes their own response and simply wants the model to clean it up or make it easier to understand. The outputs of the models were taken as is with no editing. After generating the positive examples, we passed all of the machine-generated and human-generated examples through the detectors. RADAR, Wild, and T5Sentinel all return a percentage probability for each class, and GPTID, Fastdetectgpt, Binoculars, and LogRank return a value representing their score. We do not use any thresholds and take the scores as is for AUROC and TPR@FPR metrics.

\begin{table}[t]
\centering
\resizebox{.99\linewidth}{!}{%
\begin{tabular}{lcccc}
\toprule
\textbf{Detector} & \textbf{TPR@0.01} & \textbf{TPR@0.05} & \textbf{TPR@0.1} & \textbf{AUROC} \\ \midrule

{\textbf{Radar}} & 0.05 & 0.15 & 0.27 & 0.6009 \\
{\textbf{Fast-DetectGPT}} & 0.49 & 0.61 & 0.68 & 0.8405 \\
{\textbf{Wild}} & 0.11 & 0.19 & 0.29 & 0.6841 \\
{\textbf{PHD}} & 0.08 & 0.23 & 0.37 & 0.6790 \\ 
{\textbf{LogRank}} & 0.09 & 0.40 & 0.50 & 0.7763 \\ 
{\textbf{T5Sentinel}} & 0.03 & 0.09 & 0.14 & 0.5179 \\
{\textbf{Binoculars}} & 0.58 & 0.67 & 0.72 & 0.8485\\
\bottomrule
\end{tabular}
}
\caption{Performance of different detectors across the entire dataset.}
\label{tab:detector_performance}
\end{table}

%% file: sections/05_results.tex
\begin{figure*}[htbp]
    \centering
    \begin{subfigure}[t]{0.4\textwidth}
        \centering
        \includegraphics[width=\textwidth]{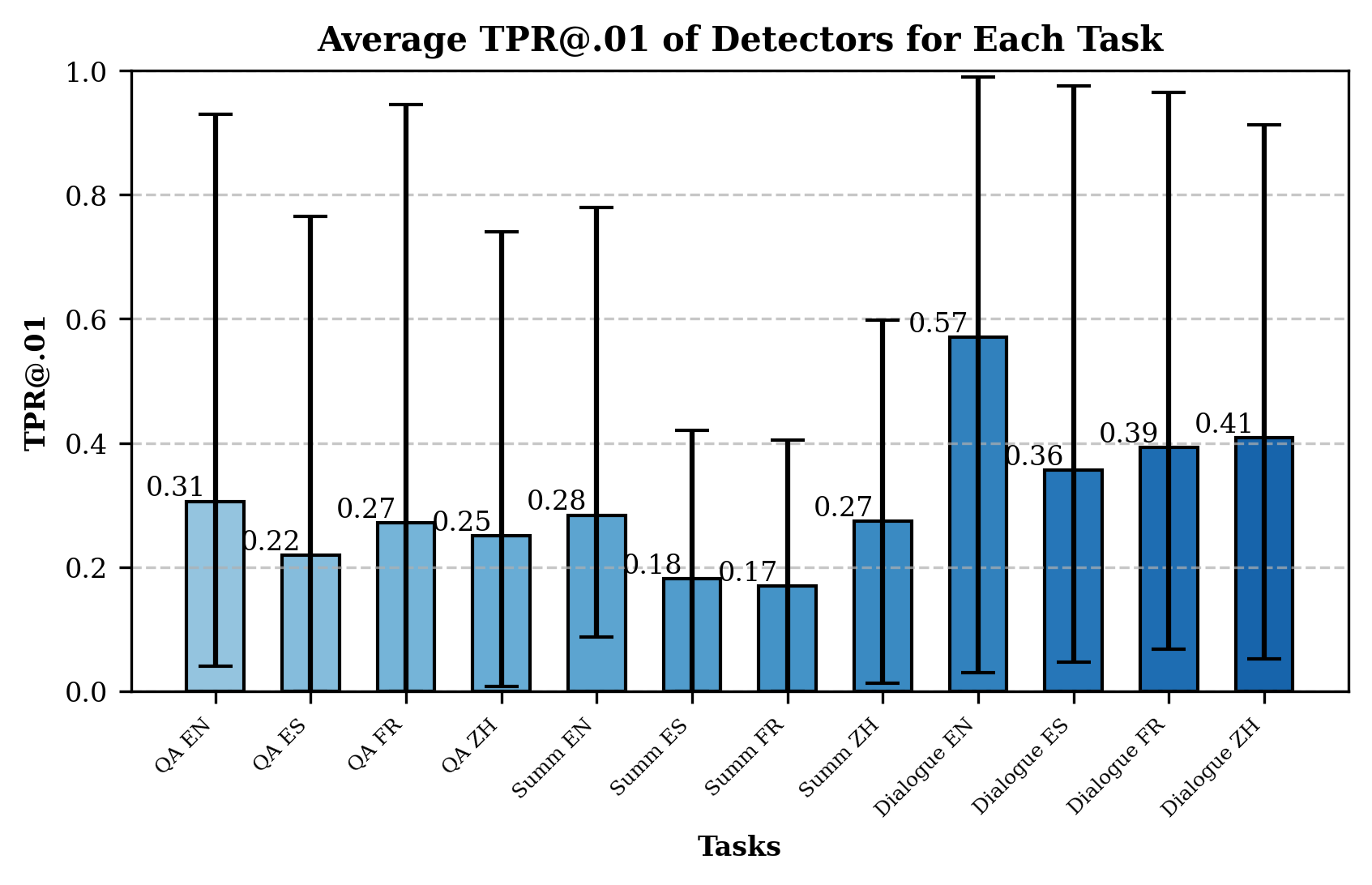}
        \caption{Average TPR@0.01 results for multilingual tasks with normal prompting across all detectors.}
        \label{fig:multi_tasks_base_tpr}
    \end{subfigure}
    \quad 
    \begin{subfigure}[t]{0.4\textwidth}
        \centering
        \includegraphics[width=\textwidth]{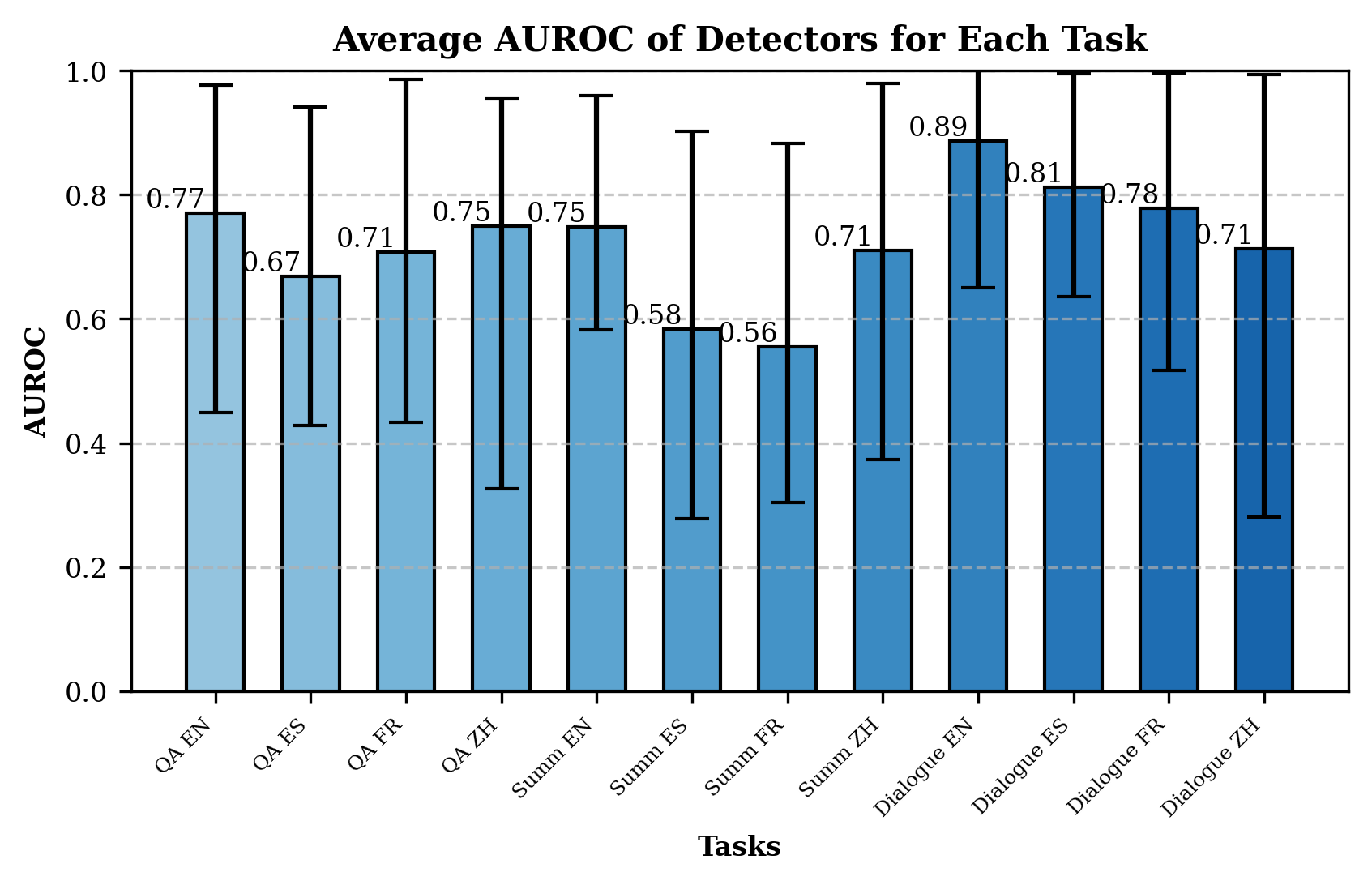}
        \caption{Average AUROC results for multilingual tasks with normal prompting across all detectors.}
        \label{fig:multi_tasks_base_auroc}
    \end{subfigure}
    
    \vspace{5mm} 
    
    \begin{subfigure}[t]{0.4\textwidth}
        \centering
        \includegraphics[width=\textwidth]{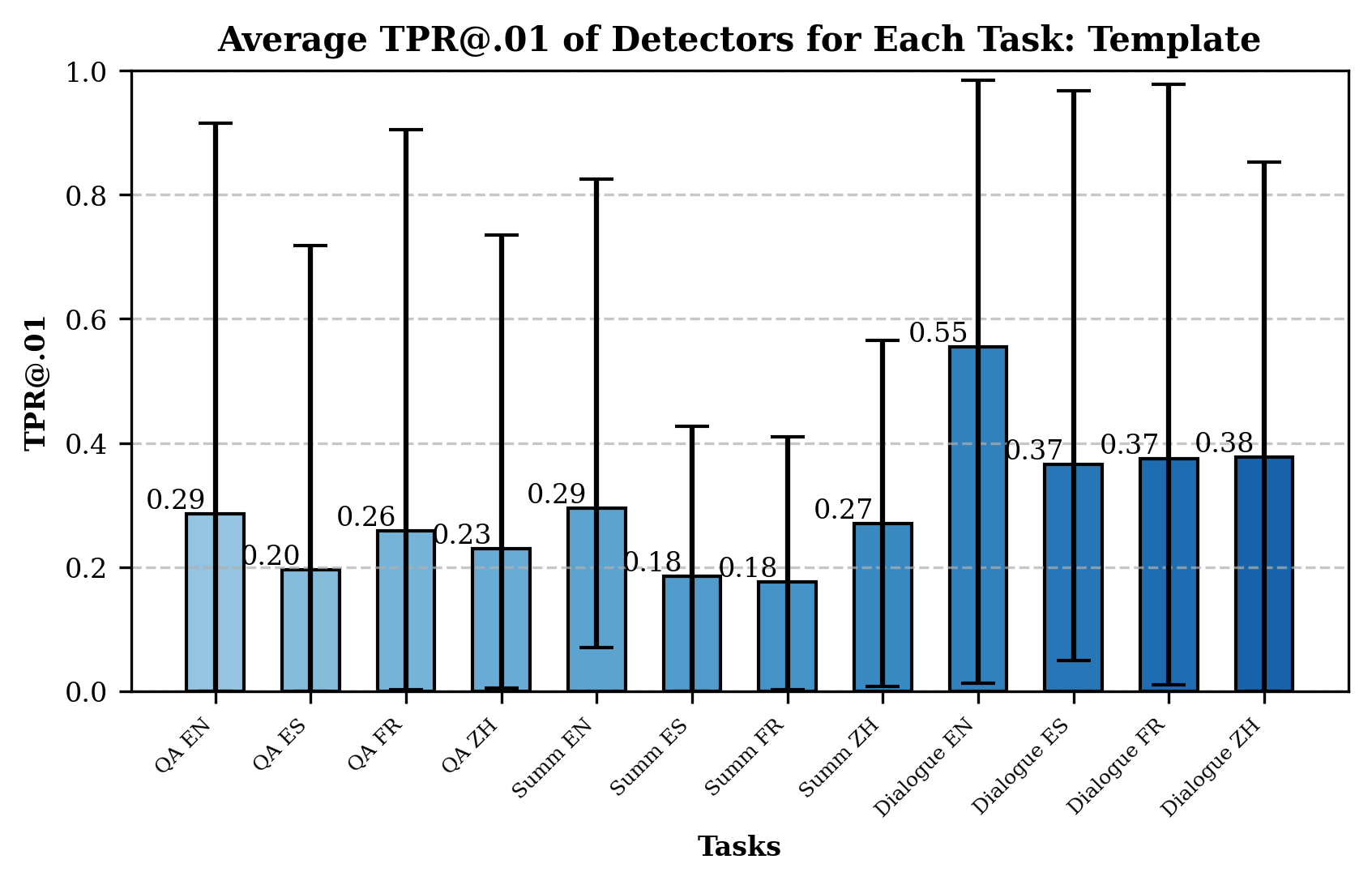}
        \caption{Average TPR@0.01 results for multilingual tasks with template prompting across all detectors.}
        \label{fig:multi_tasks_template}
    \end{subfigure}
    \quad 
    \begin{subfigure}[t]{0.4\textwidth}
        \centering
        \includegraphics[width=\textwidth]{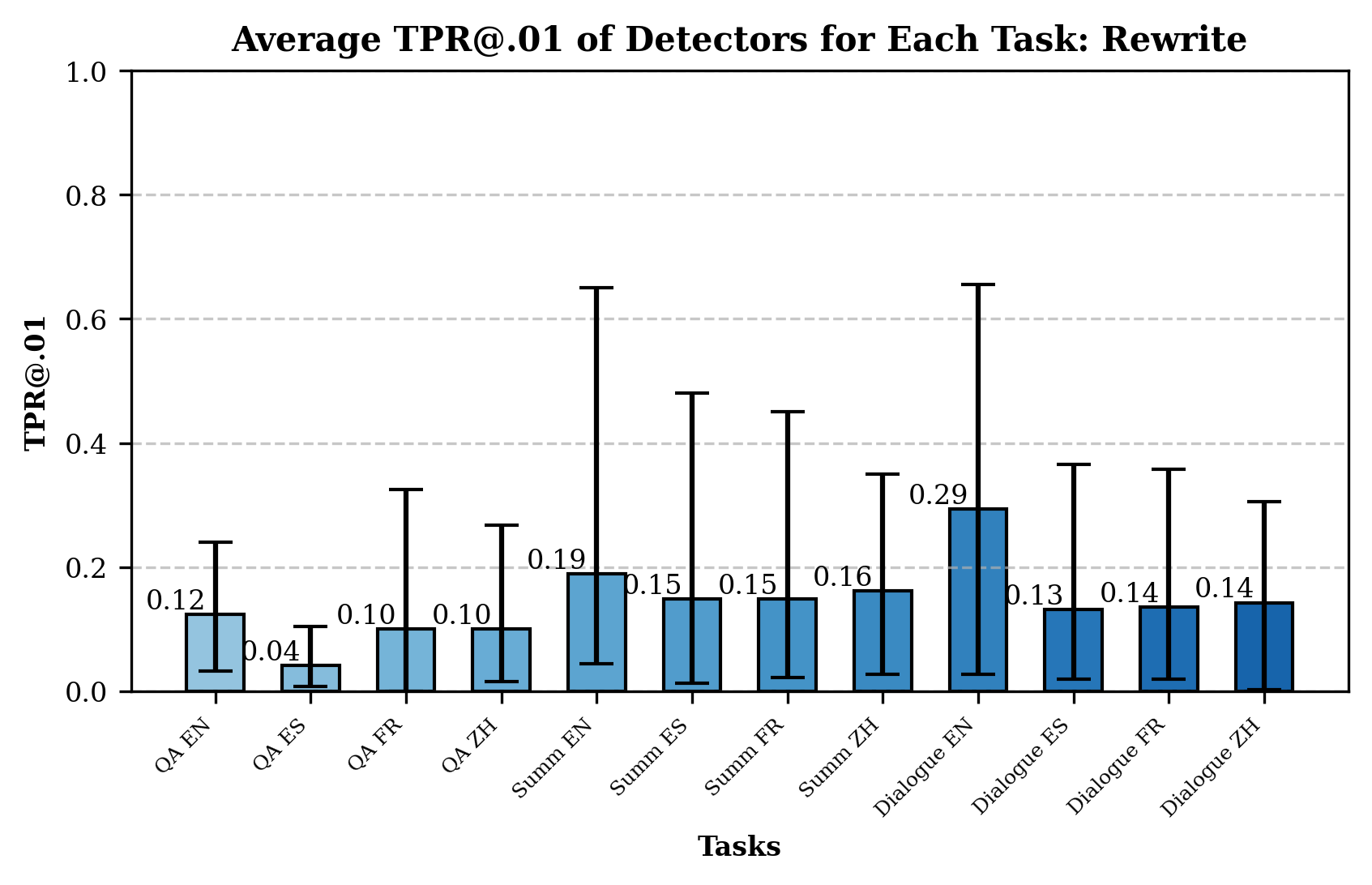}
        \caption{Average TPR@0.01 results for multilingual tasks with rewrite prompting across all detectors.}
        \label{fig:multi_tasks_rewrite}
    \end{subfigure}
    
    \caption{Comparison of average AUROC results for multilingual tasks across all detectors using different normal prompting and average TPR@0.01 across all detectors using normal, template, and rewrite prompting. Error bars show maximum and minimum performance across detectors.}
    \label{fig:multi_tasks}
\end{figure*}

\begin{figure*}[htbp]
    \centering
    \begin{subfigure}[t]{0.4\textwidth}
        \centering
        \includegraphics[width=\textwidth]{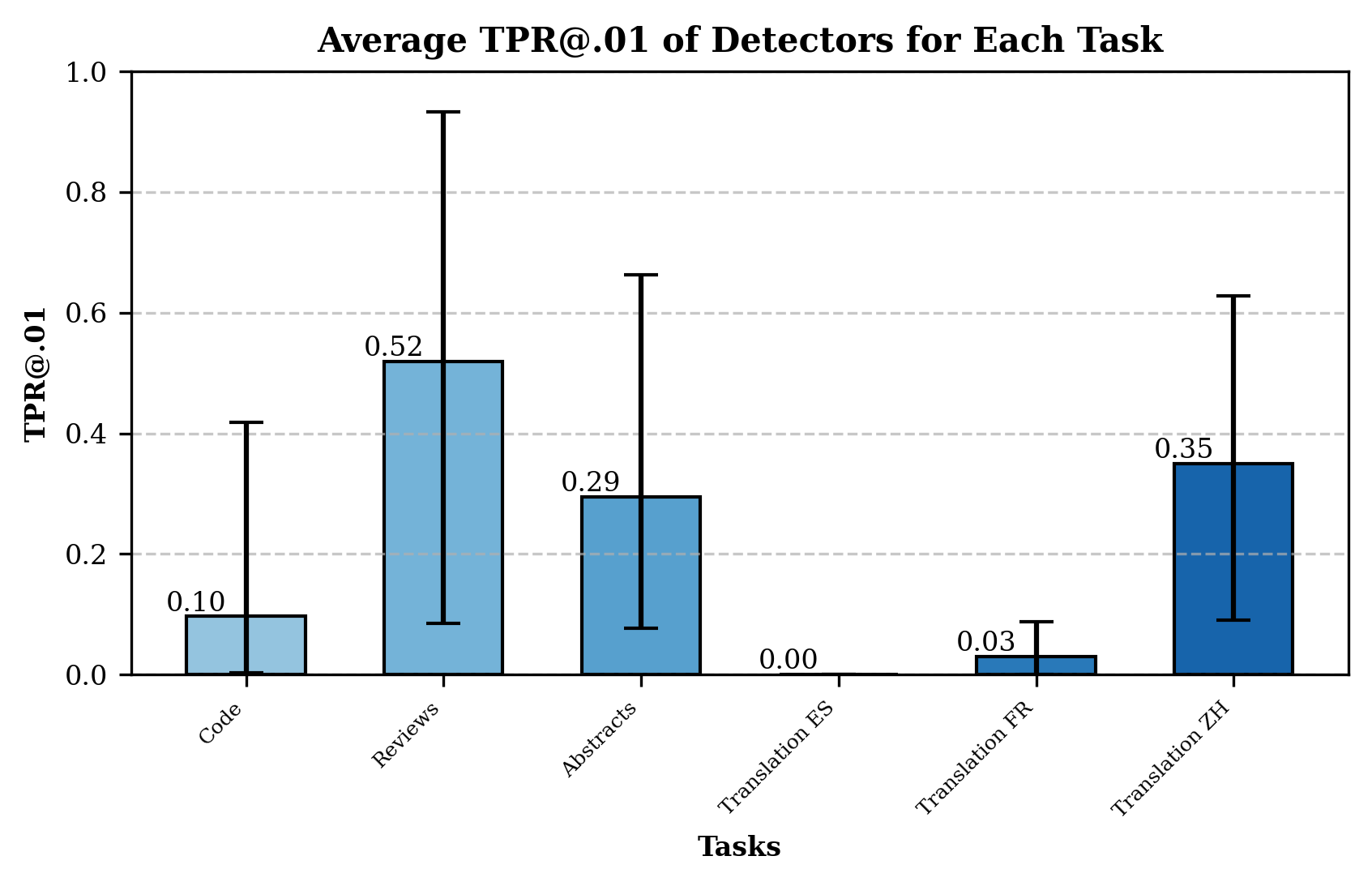}
        \caption{Average TPR@0.01 results for English tasks with normal prompting across all detectors.}
        \label{fig:other_tasks_base_tpr}
    \end{subfigure}
    \quad
    \begin{subfigure}[t]{0.4\textwidth}
        \centering
        \includegraphics[width=\textwidth]{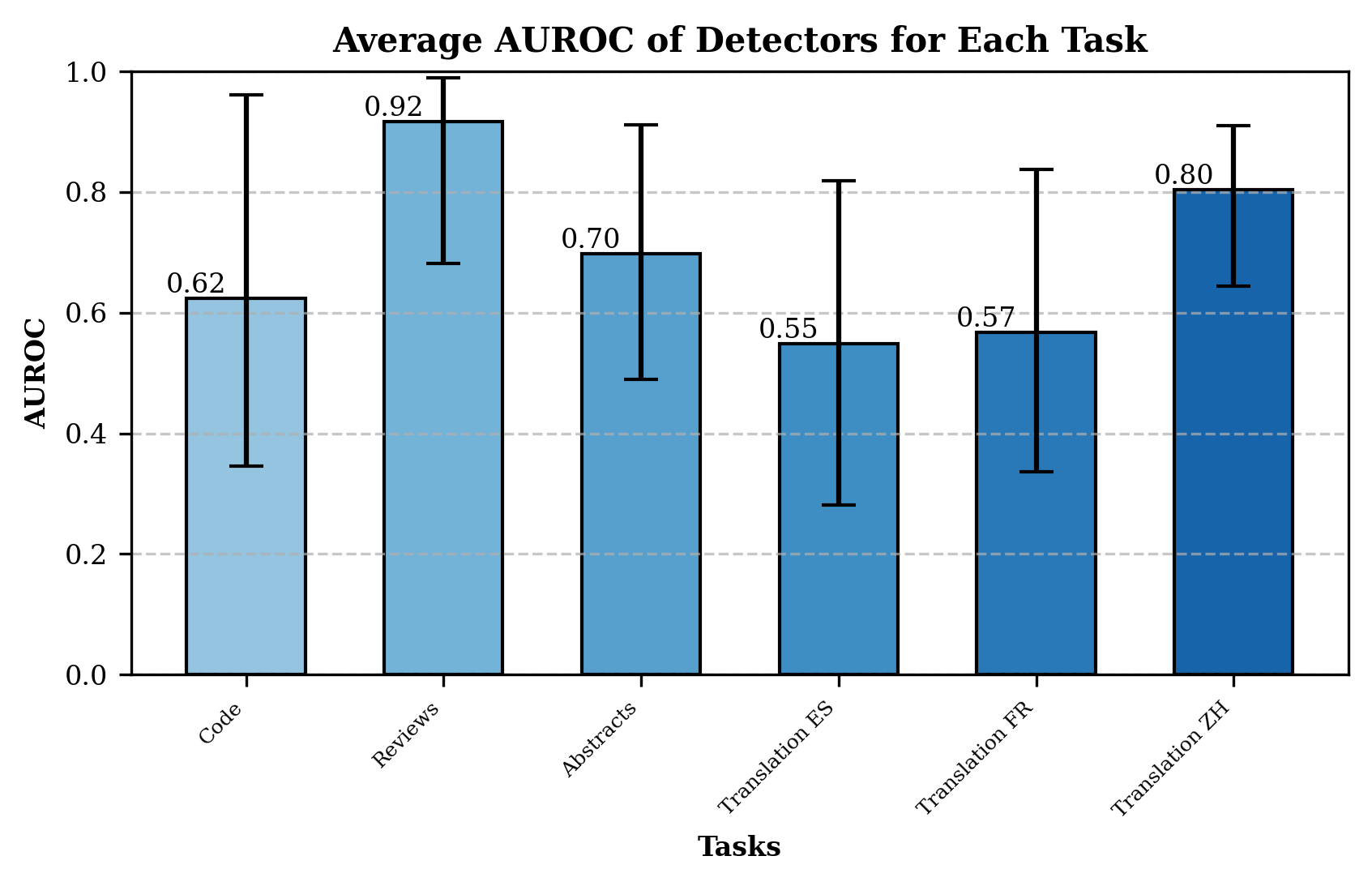}
        \caption{Average AUROC results for English tasks with normal prompting across all detectors.}
        \label{fig:other_tasks_base_auroc}
    \end{subfigure}
    
    \vspace{5mm} 
    
    \begin{subfigure}[t]{0.4\textwidth}
        \centering
        \includegraphics[width=\textwidth]{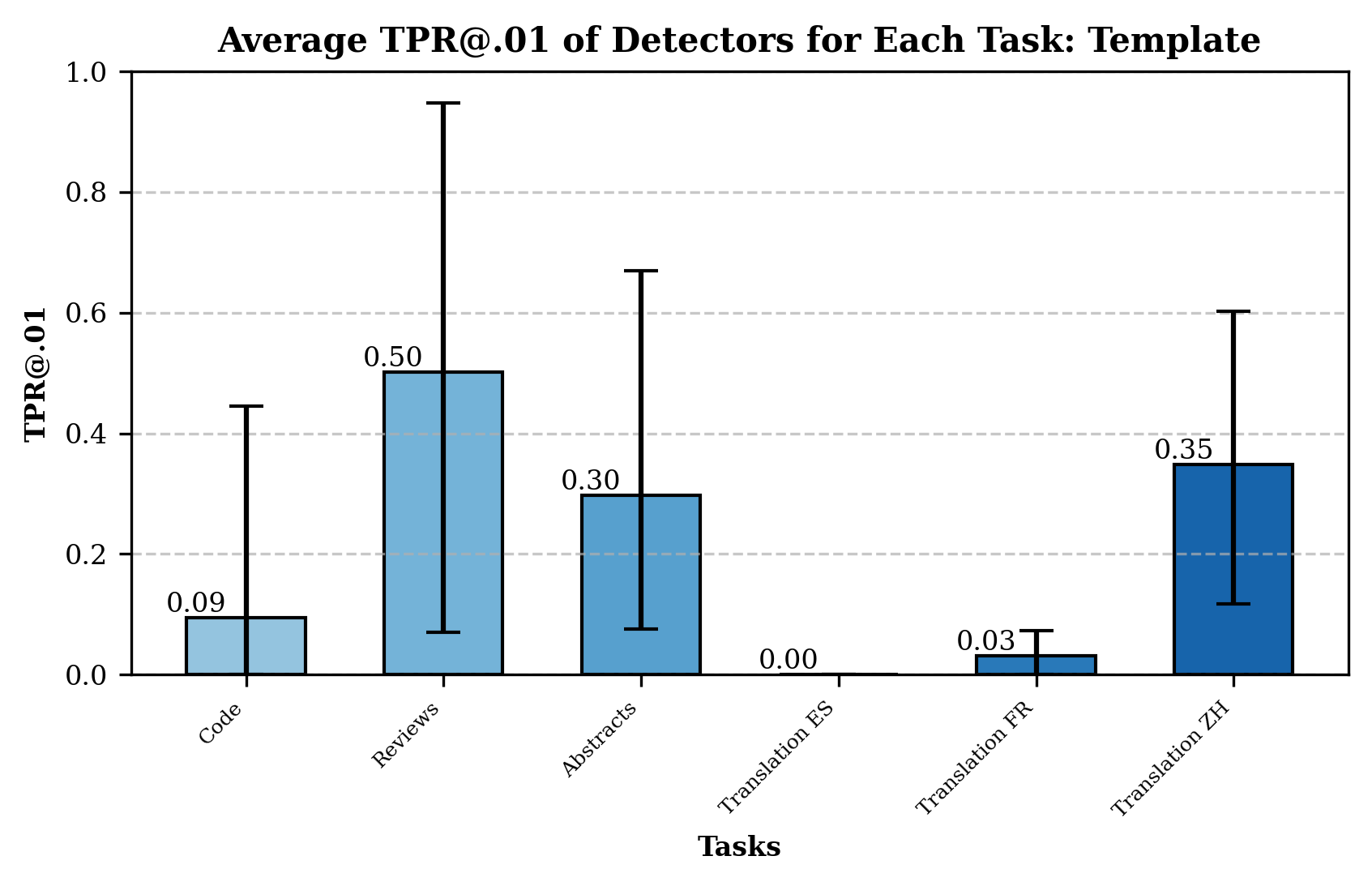}
        \caption{Average TPR@0.01 results for English tasks with template prompting across all detectors.}
        \label{fig:other_tasks_template}
    \end{subfigure}
    \quad 
    \begin{subfigure}[t]{0.4\textwidth}
        \centering
        \includegraphics[width=\textwidth]{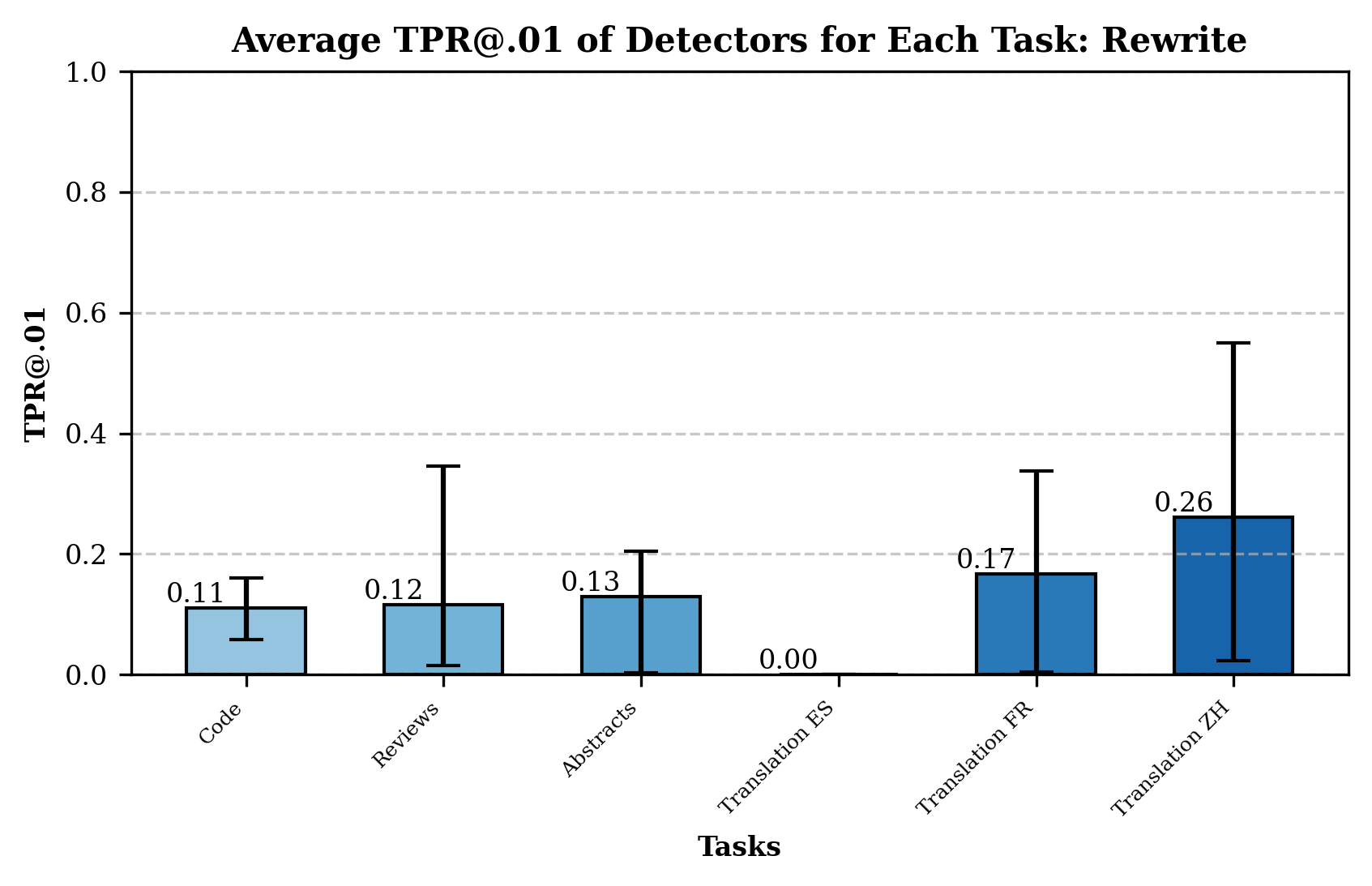}
        \caption{Average TPR@0.01 results for English tasks with rewrite prompting across all detectors.}
        \label{fig:other_tasks_rewrite}
    \end{subfigure}
    
    \caption{Comparison of average AUROC results for English tasks across all detectors using different normal prompting and average TPR@0.01 across all detectors using normal, template, and rewrite prompting. Error bars show maximum and minimum performance across detectors.}
    \label{fig:other_tasks}
\end{figure*}

Table \ref{tab:detector_performance} shows the overall performance of each detector across the entire dataset. In this section, we break down the performance of each detector across tasks, languages, and prompt techniques.

\subsection{Plain Prompting}
We evaluate the AUROC and TPR at 0.01 FPR for machine-generated texts from direct prompting using identical prompts as human written texts. A simple prompt was employed to ensure the generated text was in the correct format and language for the multilingual tasks.

Figures \ref{fig:multi_tasks_base_tpr} and \ref{fig:multi_tasks_base_auroc} show the results for the multilingual tasks and \ref{fig:other_tasks_base_tpr} and \ref{fig:other_tasks_base_auroc} show the results for the only English tasks. The results broken down by detector are shown in Appendix \ref{sec:appendix_detectors}. A significant difference is observed in detector performance across languages and tasks, particularly in the multilingual setting as well as across detectors. In the TPR@.01 setting, the difference between the best detector and worst detector is greater than $0.95$. Across all detectors we generally see strong results in the English tasks, while the performance drops off in the non-English tasks. In most detectors, in all tasks, they struggle to maintain a strong TPR rate at an FPR rate of 0.01.


For the English-only tasks, most detectors show improved performance in the AUROC, while the TPR@0.01 stays quite low. Despite expectations that the translation domain would be the most challenging due to lower entropy in translated texts, detectors performed reasonably well from the AUROC perspective. The TPR@0.01 graph highlights ongoing challenges in maintaining low false positive rates.


\subsection{Adversarial Prompting}

Figure \ref{fig:multi_tasks_template} shows the results on the multilingual tasks where the model was instructed to be "as human as possible." Interestingly, this request had little effect on performance. In the few instances where changes occurred, scores generally increased, suggesting that asking the model to "sound human" may have made its output easier to detect. This aligns with expectations, as large language models are already trained on predominantly human-written texts, and generating more conversational output can make detection more straightforward, as evidenced in dialogue generation tasks.

On the English tasks, as shown in figure, \ref{fig:other_tasks_template}, the results were similarly unaffected by the human-like request, with some slight score increases where changes were observed. This is especially expected in domains such as reviews, code, and abstracts, which follow specific writing conventions, while tasks like question answering and dialogue generation exhibit more variability and creativity.

\begin{table*}[htbp]
\resizebox{\textwidth}{!}{%
\begin{tabular}{lccccccccccccccc}
\toprule
\footnotesize\multirow{2}{*}{\textbf{Model}}   & \multicolumn{2}{c}{\footnotesize \textbf{Code}} & \multicolumn{2}{c}{\footnotesize \textbf{Reviews}} & \multicolumn{2}{c}{\footnotesize \textbf {Abstract}} & \multicolumn{2}{c}{\footnotesize \textbf{QA}} & \multicolumn{2}{c}{\footnotesize \textbf{Summ}} & \multicolumn{2}{c}{\footnotesize \textbf{Dialogue}} & \multicolumn{2}{c}{\footnotesize \textbf{Trans.}} & \footnotesize \multirow{2}{*}{\textbf{Arena Score}} \\
& \footnotesize TPR & \footnotesize AUC   & \footnotesize TPR   &  \footnotesize AUC&  \footnotesize TPR    &  \footnotesize AUC   & \footnotesize TPR   &\footnotesize AUC   & \footnotesize TPR  & \footnotesize AUC   & \footnotesize TPR  & \footnotesize AUC&\footnotesize TPR &\footnotesize AUC  &  \\ \midrule
{\footnotesize \textbf{GPT-4o}} & \footnotesize 0.02 & \footnotesize 0.55 & \footnotesize 0.28 & \footnotesize 0.63 & \footnotesize 0.04 & \footnotesize 0.53 & \footnotesize 0.05 & \footnotesize 0.54 & \footnotesize 0.05 & \footnotesize 0.50 & \footnotesize 0.03 & \footnotesize 0.58 & \footnotesize 0.03 & \footnotesize 0.52 & \footnotesize 1339 \\
{\footnotesize \textbf{Llama-3}} & \footnotesize 0.06 & \footnotesize 0.56 & \footnotesize 0.28 & \footnotesize 0.67 & \footnotesize 0.21 & \footnotesize 0.64 & \footnotesize 0.12 & \footnotesize 0.60 & \footnotesize 0.07 & \footnotesize 0.57 & \footnotesize 0.08 & \footnotesize 0.60 & \footnotesize 0.09 & \footnotesize 0.56 & \footnotesize 1152\\
{\footnotesize \textbf{Mistral}} & \footnotesize 0.02 & \footnotesize 0.54 & \footnotesize 0.28 & \footnotesize 0.65 & \footnotesize 0.04 & \footnotesize 0.54 & \footnotesize 0.10 & \footnotesize 0.58 & \footnotesize 0.04 & \footnotesize 0.51 & \footnotesize 0.06 & \footnotesize 0.59 & \footnotesize 0.05 & \footnotesize 0.54 & \footnotesize 1072\\
{\footnotesize \textbf{Phi-3}} & \footnotesize 0.04 & \footnotesize 0.57 & \footnotesize 0.24 & \footnotesize 0.62 & \footnotesize 0.13 & \footnotesize 0.58 & \footnotesize 0.08 & \footnotesize 0.58 & \footnotesize 0.12 & \footnotesize 0.58 & \footnotesize 0.13 & \footnotesize 0.63 & \footnotesize 0.08 & \footnotesize 0.56 & \footnotesize 1066\\
\bottomrule
\end{tabular}%
}
\caption{Model performance (AUROC and TPR@0.01) across tasks compared with model generation quality. The Chatbot Arena score is utilized to measure the quality of a model. The higher scores do not correlate with lower detectability of generated content.}
\label{tab:modelperf}
\end{table*}

\subsection{Rewriting}
Finally, we show the results for the rewriting prompt for the multilingual tasks in figure \ref{fig:multi_tasks_rewrite} and for the English tasks in figure \ref{fig:other_tasks_rewrite}. We observe a notable decrease in TPR@0.01 performance for detectors that previously performed well leading to a drop in the average performance in most tasks. Some of the lower performing did see an increase in performance which is why the average performance in the Code and French Translation tasks are slightly higher. Despite these shifts, the relative performance across tasks remains consistent, indicating an inherent variability in detectability based on the type of task and language.


\begin{figure}[t]
    \centering
    \includegraphics[width=.99\linewidth]{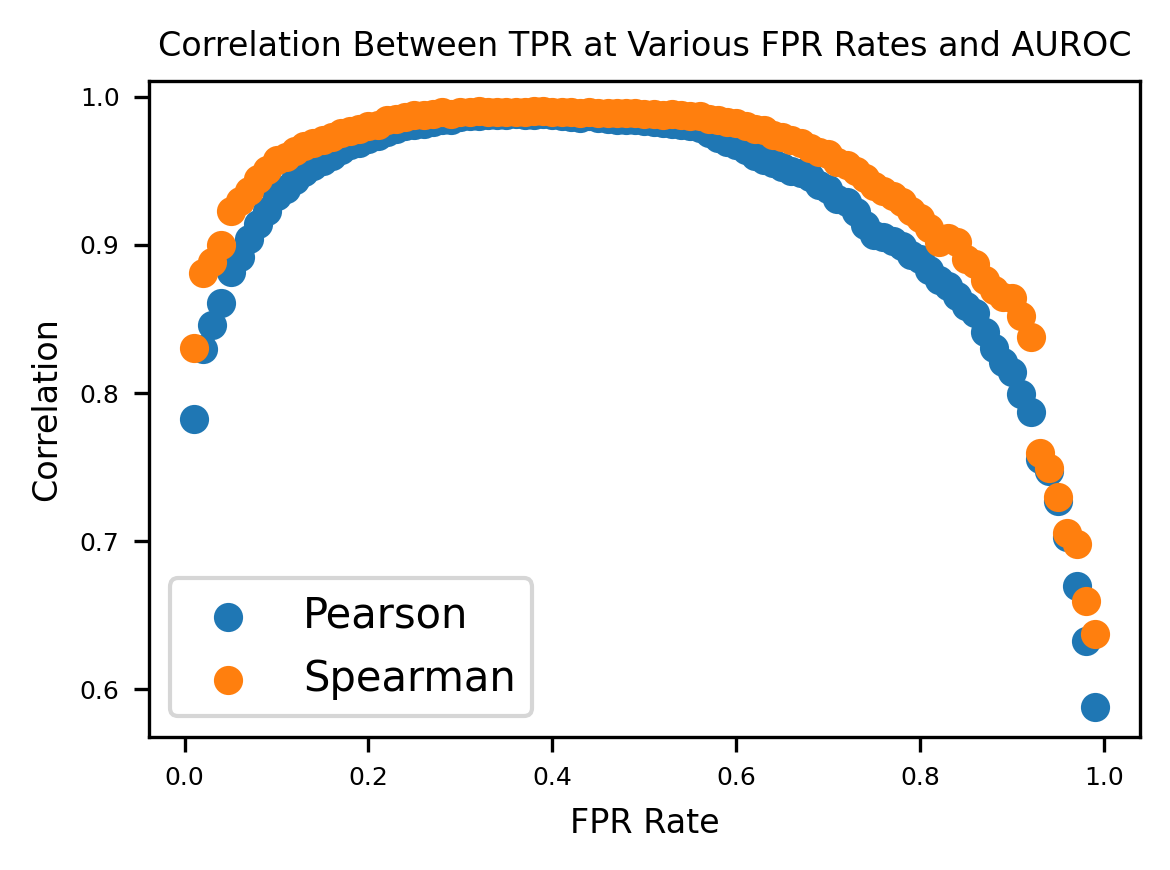} 
    \caption{Correlations between the TPR at various FPR rates and the overall AUROC score. AUROC score is more representative of the middle FPR rates, while this detection task is more concerned with the lower end of FPR.}
    \label{fig:corr}
\end{figure}

\subsection{TPR@FPR vs AUROC}

In this paper, we utilize both the AUROC and TPR@FPR metrics. However, we also argue that TPR at a low FPR is a much more important metric for this detection task. Figure \ref{fig:corr} shows the correlation between TPR scores at various FPR rates and the AUROC score for all tasks, detectors, and models used in this research. The AUROC correlates much higher with FPR rates in the 0.4 to 0.6 range and much lower with FPR rates at the edges, less than 0.2 and greater than 0.8. While the 0.75 is still a reasonable correlation value, the AUROC is still much more representative of the middle FPR's while we are really concerned with the lower FPR's for this task. This is why we report the TPR@0.01, which is much more representative of the applicability of a detector than the AUROC.

\subsection{Output Quality and Detection}

Measuring the quality of LLM outputs, especially in creative tasks, remains challenging, making it difficult to determine if higher-quality outputs are harder to detect. Table \ref{tab:modelperf} compares various models' performance scores and rankings from Chatbot Arena \citep{chiang2024chatbot}, allowing us to explore if output quality affects detectability. The data shows little difference in detectability across models of varying quality, with AUROC and TPR@0.01 scores remaining consistent. This suggests that output quality does not significantly impact the difficulty of detection, though further research is needed for a fuller understanding.

\begin{figure}[t]
    \centering
    \includegraphics[width=.99\linewidth]{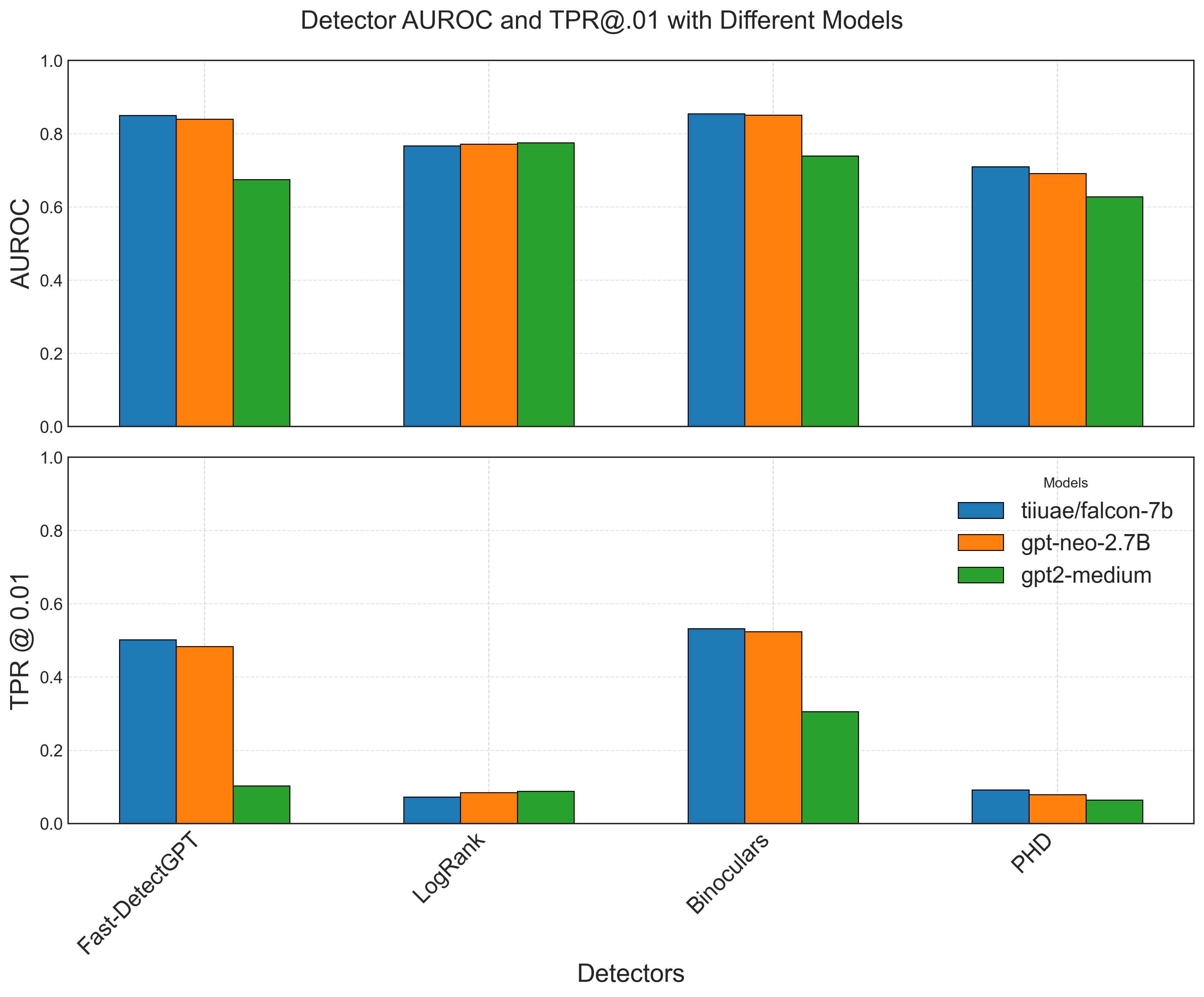} 
    \caption{AUROC and TPR@0.01 for each zero-shot method using various underlying models. Only Fast-DetectGPT and Binoculars show a significant change in performance with GPT2-Medium.}
    \label{fig:zeroshot}
\end{figure}

\subsection{Impact of Model on Zero-shot Methods}
Each zero-shot method used in this paper has an underlying model that assists in the detection process. In this paper we consider the model chosen by the respective authors of each detector to be a part of the detector itself. However, we also swapped out each model to directly compare the statistical methods themselves, removing any impact from a specific model on a detector.

Figure \ref{fig:zeroshot} shows the results of running each detector across the entire dataset with three different models. There is generally not much of a difference in the ability of a detection method when changing the underlying model. Fast-DetectGPT and Binoculars show a small change in AUROC and a larger change in TPR@0.01 when using the gpt2-medium model \citep{radford2019language}. Gpt2-medium is the oldest model of the three, which likely results in its output logits being different than the generation models more often. This provides some evidence that these zero-shot methods will require updated underlying models to remain successful on more advanced generation models, but more research would need to be conducted.

%% file: sections/07_conclusion.tex
This study evaluates seven advanced detectors across seven tasks and four languages, revealing notable inconsistencies in their detection capabilities. We also examined three different prompting strategies and their impact on detectability, finding that requests for more "human-like" output do not make the text harder to detect, while rewritten human content proves more difficult to identify. 

The detection results for both the Translation task and Rewrite prompt are generally lower than the average detectability for other machine generated text. This encourages a discussion about whether this type of text should be detected as machine generated or not. The text may have been machine generated, but it is heavily influenced by human generated text. Specifically in the translation case, the text should match the source text in another language. In the Rewrite case, the model is not generating any new ideas, just improving the readability of the text. It is clear from the results that these cases are harder to detect than when a model has a more open-ended generation. It is likely worth differentiating between these two cases which we leave to future work.

Additionally, this research highlights the limitations of relying on the AUROC metric for assessing machine-generated content detectors. Our findings emphasize the need for robust evaluation methods to develop more reliable detection techniques. The study underscores the challenges in detecting machine-generated text, particularly when human written text was only modified by a language model, and advocates for TPR@FPR as the preferred evaluation metric to better capture detector performance.

%% file: sections/09_limitations.tex
A limitation of this method is the settings in which the human data was collected may vary from the settings in which these detectors will be used. Additionally, some of the datasets we used had collected their data from the internet which raises a concern that some of that data is not completely human generated. This is a challenge that all future detectors will also struggle with when training and evaluating. These results pose the risk of emboldening users to use AI generated content when they otherwise should not because they know detectors cannot be confidently trusted. However, acknowledging this is important to encouraging research into new detection methods and improving current methods.

%% file: sections/08_appendix.tex
This section contains results for detections by models and tasks, the prompts used for plain prompting, and the results by detector.

\subsection{Results by Model}
The following tables show the results for each detector by generation model and task. As discussed in the paper, there is not a significant difference in detectability of a text by the model that generated that text. More specifically, a higher quality model like GPT-4o is not noticeably harder to detect than a lower quality model like Phi-3. The differences in detectability are more obvious across tasks than generation models.

\begin{table}[htbp]
\small
\centering
\begin{tabular}{llcc}
\toprule
\textbf{Model} & \textbf{Detector} & \textbf{TPR@.01} & \textbf{AUROC} \\ \midrule
\multirow{7}{*}{\textbf{GPT-4o}} & Binoculars & 0.14 & 0.8739 \\
 & Fast-DetectGPT & 0.07 & 0.8223 \\
 & LogRank & 0.00 & 0.5722 \\
 & PHD & 0.00 & 0.3976 \\
 & Radar & 0.00 & 0.6594 \\
 & T5Sentinel & 0.01 & 0.4604 \\
 & Wild & 0.00 & 0.4724 \\
\midrule
\multirow{7}{*}{\textbf{Llama-3}} & Binoculars & 0.45 & 0.9474 \\
 & Fast-DetectGPT & 0.16 & 0.8211 \\
 & LogRank & 0.00 & 0.6060 \\
 & PHD & 0.04 & 0.4940 \\
 & Radar & 0.05 & 0.8088 \\
 & T5Sentinel & 0.09 & 0.5358 \\
 & Wild & 0.06 & 0.6231 \\
\midrule
\multirow{7}{*}{\textbf{Mistral}} & Binoculars & 0.39 & 0.9681 \\
 & Fast-DetectGPT & 0.16 & 0.9029 \\
 & LogRank & 0.06 & 0.4552 \\
 & PHD & 0.06 & 0.3104 \\
 & Radar & 0.00 & 0.5787 \\
 & T5Sentinel & 0.02 & 0.3315 \\
 & Wild & 0.07 & 0.5412 \\
\midrule
\multirow{7}{*}{\textbf{Phi-3}} & Binoculars & 0.35 & 0.7825 \\
 & Fast-DetectGPT & 0.21 & 0.8076 \\
 & LogRank & 0.08 & 0.5470 \\
 & PHD & 0.14 & 0.4657 \\
 & Radar & 0.06 & 0.7700 \\
 & T5Sentinel & 0.03 & 0.5209 \\
 & Wild & 0.12 & 0.5779 \\
\bottomrule
\end{tabular}
\caption{Code}
\label{tab:code_results}
\end{table}

\begin{table}[htbp]
\small
\centering
\begin{tabular}{llcc}
\toprule
\textbf{Model} & \textbf{Detector} & \textbf{TPR@.01} & \textbf{AUROC} \\ \midrule
\multirow{7}{*}{\textbf{GPT-4o}} & Binoculars & 0.47 & 0.8676 \\
 & Fast-DetectGPT & 0.35 & 0.8918 \\
 & LogRank & 0.01 & 0.5972 \\
 & PHD & 0.01 & 0.4758 \\
 & Radar & 0.02 & 0.3474 \\
 & T5Sentinel & 0.01 & 0.4909 \\
 & Wild & 0.02 & 0.5088 \\
\midrule
\multirow{7}{*}{\textbf{Llama-3}} & Binoculars & 0.66 & 0.9501 \\
 & Fast-DetectGPT & 0.68 & 0.9465 \\
 & LogRank & 0.03 & 0.7816 \\
 & PHD & 0.02 & 0.6637 \\
 & Radar & 0.13 & 0.6581 \\
 & T5Sentinel & 0.03 & 0.4995 \\
 & Wild & 0.06 & 0.6024 \\
\midrule
\multirow{7}{*}{\textbf{Mistral}} & Binoculars & 0.56 & 0.8954 \\
 & Fast-DetectGPT & 0.57 & 0.8852 \\
 & LogRank & 0.04 & 0.7052 \\
 & PHD & 0.04 & 0.6261 \\
 & Radar & 0.04 & 0.5776 \\
 & T5Sentinel & 0.02 & 0.5078 \\
 & Wild & 0.04 & 0.5664 \\
\midrule
\multirow{7}{*}{\textbf{Phi-3}} & Binoculars & 0.48 & 0.8188 \\
 & Fast-DetectGPT & 0.41 & 0.8424 \\
 & LogRank & 0.13 & 0.7448 \\
 & PHD & 0.13 & 0.6553 \\
 & Radar & 0.07 & 0.5966 \\
 & T5Sentinel & 0.02 & 0.5787 \\
 & Wild & 0.07 & 0.5923 \\
\bottomrule
\end{tabular}
\caption{Question Answering}
\label{tab:qa_results}
\end{table}

\begin{table}[htbp]
\small
\centering
\begin{tabular}{llcc}
\toprule
\textbf{Model} & \textbf{Detector} & \textbf{TPR@.01} & \textbf{AUROC} \\ \midrule
\multirow{7}{*}{\textbf{GPT-4o}} & Binoculars & 0.05 & 0.6533 \\
 & Fast-DetectGPT & 0.11 & 0.6981 \\
 & LogRank & 0.23 & 0.7070 \\
 & PHD & 0.00 & 0.4922 \\
 & Radar & 0.00 & 0.2085 \\
 & T5Sentinel & 0.01 & 0.4339 \\
 & Wild & 0.11 & 0.4753 \\
\midrule
\multirow{7}{*}{\textbf{Llama-3}} & Binoculars & 0.33 & 0.8134 \\
 & Fast-DetectGPT & 0.19 & 0.7540 \\
 & LogRank & 0.61 & 0.8941 \\
 & PHD & 0.30 & 0.6642 \\
 & Radar & 0.08 & 0.6605 \\
 & T5Sentinel & 0.05 & 0.4720 \\
 & Wild & 0.15 & 0.6570 \\
\midrule
\multirow{7}{*}{\textbf{Mistral}} & Binoculars & 0.06 & 0.6011 \\
 & Fast-DetectGPT & 0.08 & 0.5862 \\
 & LogRank & 0.30 & 0.7635 \\
 & PHD & 0.01 & 0.5216 \\
 & Radar & 0.00 & 0.3428 \\
 & T5Sentinel & 0.02 & 0.4256 \\
 & Wild & 0.13 & 0.5931 \\
\midrule
\multirow{7}{*}{\textbf{Phi-3}} & Binoculars & 0.48 & 0.7501 \\
 & Fast-DetectGPT & 0.30 & 0.7135 \\
 & LogRank & 0.60 & 0.9049 \\
 & PHD & 0.33 & 0.7668 \\
 & Radar & 0.09 & 0.7492 \\
 & T5Sentinel & 0.01 & 0.3561 \\
 & Wild & 0.38 & 0.8890 \\
\bottomrule
\end{tabular}
\caption{Summarization}
\label{tab:summ_results}
\end{table}

\begin{table}[htbp]
\small
\centering
\begin{tabular}{llcc}
\toprule
\textbf{Model} & \textbf{Detector} & \textbf{TPR@.01} & \textbf{AUROC} \\ \midrule
\multirow{7}{*}{\textbf{GPT-4o}} & Binoculars & 0.68 & 0.9362 \\
 & Fast-DetectGPT & 0.52 & 0.9277 \\
 & LogRank & 0.37 & 0.7857 \\
 & PHD & 0.07 & 0.6279 \\
 & Radar & 0.03 & 0.6143 \\
 & T5Sentinel & 0.06 & 0.5045 \\
 & Wild & 0.01 & 0.5633 \\
\midrule
\multirow{7}{*}{\textbf{Llama-3}} & Binoculars & 0.80 & 0.9695 \\
 & Fast-DetectGPT & 0.67 & 0.9487 \\
 & LogRank & 0.42 & 0.8235 \\
 & PHD & 0.24 & 0.6772 \\
 & Radar & 0.14 & 0.6907 \\
 & T5Sentinel & 0.04 & 0.5265 \\
 & Wild & 0.03 & 0.6378 \\
\midrule
\multirow{7}{*}{\textbf{Mistral}} & Binoculars & 0.64 & 0.9150 \\
 & Fast-DetectGPT & 0.56 & 0.9246 \\
 & LogRank & 0.33 & 0.7827 \\
 & PHD & 0.17 & 0.6514 \\
 & Radar & 0.19 & 0.6989 \\
 & T5Sentinel & 0.02 & 0.4949 \\
 & Wild & 0.02 & 0.6044 \\
\midrule
\multirow{7}{*}{\textbf{Phi-3}} & Binoculars & 0.69 & 0.9243 \\
 & Fast-DetectGPT & 0.59 & 0.8506 \\
 & LogRank & 0.60 & 0.9206 \\
 & PHD & 0.37 & 0.7745 \\
 & Radar & 0.04 & 0.7626 \\
 & T5Sentinel & 0.11 & 0.6134 \\
 & Wild & 0.10 & 0.6505 \\
\bottomrule
\end{tabular}
\caption{Dialogue}
\label{tab:dia_results}
\end{table}

\begin{table}[htbp]
\small
\centering
\begin{tabular}{llcc}
\toprule
\textbf{Model} & \textbf{Detector} & \textbf{TPR@.01} & \textbf{AUROC} \\ \midrule
\multirow{7}{*}{\textbf{GPT-4o}} & Binoculars & 0.39 & 0.8584 \\
 & Fast-DetectGPT & 0.33 & 0.8895 \\
 & LogRank & 0.00 & 0.6474 \\
 & PHD & 0.00 & 0.2399 \\
 & Radar & 0.01 & 0.1890 \\
 & T5Sentinel & 0.00 & 0.3086 \\
 & Wild & 0.06 & 0.6178 \\
\midrule
\multirow{7}{*}{\textbf{Llama-3}} & Binoculars & 0.71 & 0.9248 \\
 & Fast-DetectGPT & 0.70 & 0.9352 \\
 & LogRank & 0.12 & 0.8505 \\
 & PHD & 0.09 & 0.5909 \\
 & Radar & 0.29 & 0.6430 \\
 & T5Sentinel & 0.16 & 0.6843 \\
 & Wild & 0.42 & 0.8185 \\
\midrule
\multirow{7}{*}{\textbf{Mistral}} & Binoculars & 0.44 & 0.8672 \\
 & Fast-DetectGPT & 0.38 & 0.8937 \\
 & LogRank & 0.01 & 0.6964 \\
 & PHD & 0.01 & 0.3262 \\
 & Radar & 0.03 & 0.1770 \\
 & T5Sentinel & 0.05 & 0.4642 \\
 & Wild & 0.04 & 0.5601 \\
\midrule
\multirow{7}{*}{\textbf{Phi-3}} & Binoculars & 0.38 & 0.5372 \\
 & Fast-DetectGPT & 0.43 & 0.7147 \\
 & LogRank & 0.30 & 0.8252 \\
 & PHD & 0.48 & 0.7344 \\
 & Radar & 0.42 & 0.8498 \\
 & T5Sentinel & 0.01 & 0.3748 \\
 & Wild & 0.49 & 0.9189 \\
\bottomrule
\end{tabular}
\caption{Abstract}
\label{tab:abs_results}
\end{table}

\begin{table}[htbp]
\small
\centering
\begin{tabular}{llcc}
\toprule
\textbf{Model} & \textbf{Detector} & \textbf{TPR@.01} & \textbf{AUROC} \\ \midrule
\multirow{7}{*}{\textbf{GPT-4o}} & Binoculars & 0.69 & 0.9247 \\
 & Fast-DetectGPT & 0.65 & 0.8847 \\
 & LogRank & 0.31 & 0.8778 \\
 & PHD & 0.00 & 0.7663 \\
 & Radar & 0.12 & 0.8791 \\
 & T5Sentinel & 0.02 & 0.5718 \\
 & Wild & 0.00 & 0.9249 \\
\midrule
\multirow{7}{*}{\textbf{Llama-3}} & Binoculars & 0.80 & 0.9662 \\
 & Fast-DetectGPT & 0.72 & 0.9305 \\
 & LogRank & 0.65 & 0.9240 \\
 & PHD & 0.47 & 0.8589 \\
 & Radar & 0.39 & 0.8883 \\
 & T5Sentinel & 0.11 & 0.5814 \\
 & Wild & 0.30 & 0.9541 \\
\midrule
\multirow{7}{*}{\textbf{Mistral}} & Binoculars & 0.79 & 0.9659 \\
 & Fast-DetectGPT & 0.74 & 0.9358 \\
 & LogRank & 0.60 & 0.9249 \\
 & PHD & 0.06 & 0.8347 \\
 & Radar & 0.37 & 0.9015 \\
 & T5Sentinel & 0.07 & 0.6376 \\
 & Wild & 0.42 & 0.9574 \\
\midrule
\multirow{7}{*}{\textbf{Phi-3}} & Binoculars & 0.70 & 0.8889 \\
 & Fast-DetectGPT & 0.49 & 0.8178 \\
 & LogRank & 0.40 & 0.8272 \\
 & PHD & 0.17 & 0.8137 \\
 & Radar & 0.47 & 0.9024 \\
 & T5Sentinel & 0.03 & 0.4166 \\
 & Wild & 0.25 & 0.9410 \\
\bottomrule
\end{tabular}
\caption{Reviews}
\label{tab:rev_results}
\end{table}

\begin{table}[htbp]
\small
\centering
\begin{tabular}{llcc}
\toprule
\textbf{Model} & \textbf{Detector} & \textbf{TPR@.01} & \textbf{AUROC} \\ \midrule
\multirow{7}{*}{\textbf{GPT-4o}} & Binoculars & 0.12 & 0.7020 \\
 & Fast-DetectGPT & 0.05 & 0.6539 \\
 & LogRank & 0.02 & 0.6059 \\
 & PHD & 0.02 & 0.5611 \\
 & Radar & 0.01 & 0.6624 \\
 & T5Sentinel & 0.01 & 0.3732 \\
 & Wild & 0.13 & 0.5788 \\
\midrule
\multirow{7}{*}{\textbf{Llama-3}} & Binoculars & 0.54 & 0.8739 \\
 & Fast-DetectGPT & 0.40 & 0.8146 \\
 & LogRank & 0.23 & 0.7281 \\
 & PHD & 0.08 & 0.6874 \\
 & Radar & 0.18 & 0.8910 \\
 & T5Sentinel & 0.06 & 0.4902 \\
 & Wild & 0.37 & 0.7462 \\
\midrule
\multirow{7}{*}{\textbf{Mistral}} & Binoculars & 0.31 & 0.7781 \\
 & Fast-DetectGPT & 0.14 & 0.7139 \\
 & LogRank & 0.05 & 0.6167 \\
 & PHD & 0.03 & 0.6115 \\
 & Radar & 0.08 & 0.8602 \\
 & T5Sentinel & 0.02 & 0.3869 \\
 & Wild & 0.19 & 0.6562 \\
\midrule
\multirow{7}{*}{\textbf{Phi-3}} & Binoculars & 0.36 & 0.7671 \\
 & Fast-DetectGPT & 0.20 & 0.6446 \\
 & LogRank & 0.09 & 0.6377 \\
 & PHD & 0.30 & 0.7130 \\
 & Radar & 0.36 & 0.9726 \\
 & T5Sentinel & 0.01 & 0.4116 \\
 & Wild & 0.45 & 0.8199 \\
\bottomrule
\end{tabular}
\caption{Translation}
\label{tab:trans_results}
\end{table}

\subsection{Plain Prompts}
Table \ref{tab:plain_prompts} shows the prompts used for each task in the plain prompting. The prompts we used were intentionally very simple and not overly instructive. This is because we wanted to replicate a realistic scenario of an average person prompting a language model. We performed small ablations on these prompts and found no difference in detectability. 
\begin{table*}[!htbp]
\centering
\resizebox{0.8\linewidth}{!}{
\begin{tabular}{lp{10cm}}
\toprule
\textbf{Task} & \textbf{Prompt} \\
\midrule
Code & You are a helpful code assistant that can teach a junior developer how to code. Your language of choice is Python. Don't explain the code, just generate the code block itself. \\
\midrule
Question Answering & You are a helpful question answering assistant who will answer a single question as completely as possible given the information in the question. Do NOT use any markdown, bullet, or numbered list formatting. The assistant will use ONLY paragraph formatting. **Respond only in \{language\}** \\
\midrule
Summarization & You are a helpful summarization assistant that will summarize a given article. Provide only the summarization in paragraph formatting. Do not introduce the summary. **Respond in \{language\}**\\
\midrule
Dialogue & You are a helpful dialogue generation assistant that will generate a dialogue between people given a short paragraph describing the people involved. Provide only the dialogue. Do not introduce the dialogue. **Respond in \{language\}** \\
\midrule
Abstract Writing & You are a helpful abstract writing assistant. You will write an abstract given the content of a paper. Do not provide any other text. You will only provide an abstract. \\
\midrule
Review Writing & You are a helpful conference paper review assistant. Please provide a detailed review of the following paper, including its strengths, weaknesses, and suggestions for improvement. \\
\midrule
Translation & You are a helpful translation assistant that will translate a given text into English. Provide only the translation and nothing else. \\
\midrule
Rewriting & You are a helpful writing assistant. Rewrite the following text to improve clarity and professionalism. Do not provide any other text. Only provide the rewritten text.\\
\bottomrule
\end{tabular}
}
\caption{The table shows the prompts used in the plain prompting. For GPT, these were used as system prompts, and for huggingface models they were prepended to the questions.}
\label{tab:plain_prompts}
\end{table*}

\clearpage
\onecolumn
\subsection{Results by Detector}\label{sec:appendix_detectors}

This section shows the numerical value of each detector on each task. In the paper we display graphs representing most of these values but show all of the numbers here for reference. The TPR@.01 and AUROC change significantly across tasks for every detector signifying that these detectors are not equally capable of detecting all types of machine generated text.

\begin{table*}[!htbp]
\resizebox{\textwidth}{!}{%
\begin{tabular}{lcccccccccccc}
\toprule
  & \multicolumn{2}{c}{\footnotesize \textbf{Code}} & \multicolumn{2}{c}{\footnotesize \textbf{Reviews}} & \multicolumn{2}{c}{\footnotesize \textbf{Abstract}} & \multicolumn{2}{c}{\footnotesize \textbf{Translation ES}} & \multicolumn{2}{c}{\footnotesize \textbf{Translation FR}} & \multicolumn{2}{c}{\footnotesize \textbf{Translation ZH}} \\
  & \footnotesize TPR & \footnotesize AUC & \footnotesize TPR & \footnotesize AUC & \footnotesize TPR & \footnotesize AUC & \footnotesize TPR & \footnotesize AUC & \footnotesize TPR & \footnotesize AUC & \footnotesize TPR & \footnotesize AUC \\
\midrule
{\footnotesize \textbf{Radar}} & \footnotesize 0.0258 & \footnotesize 0.7042 & \footnotesize 0.3358 & \footnotesize 0.8928 & \footnotesize 0.1858 & \footnotesize 0.4647 & \footnotesize 0.1806 & \footnotesize 0.8528 & \footnotesize 0.1475 & \footnotesize 0.8565 & \footnotesize 0.3625 & \footnotesize 0.8307 \\
{\footnotesize \textbf{Fast-DetectGPT}} & \footnotesize 0.1508 & \footnotesize 0.8385 & \footnotesize 0.6475 & \footnotesize 0.8922 & \footnotesize 0.4608 & \footnotesize 0.8583 & \footnotesize 0.1782 & \footnotesize 0.6337 & \footnotesize 0.0817 & \footnotesize 0.6078 & \footnotesize 0.3333 & \footnotesize 0.7953 \\
{\footnotesize \textbf{Wild}} & \footnotesize 0.0608 & \footnotesize 0.5537 & \footnotesize 0.2425 & \footnotesize 0.9443 & \footnotesize 0.2517 & \footnotesize 0.7288 & \footnotesize 0.1759 & \footnotesize 0.7433 & \footnotesize 0.0875 & \footnotesize 0.5455 & \footnotesize 0.5358 & \footnotesize 0.9049 \\
{\footnotesize \textbf{PHD}} & \footnotesize 0.0625 & \footnotesize 0.4169 & \footnotesize 0.1758 & \footnotesize 0.8184 & \footnotesize 0.1467 & \footnotesize 0.4729 & \footnotesize 0.0556 & \footnotesize 0.5264 & \footnotesize 0.0075 & \footnotesize 0.6328 & \footnotesize 0.2283 & \footnotesize 0.7915 \\
{\footnotesize \textbf{LogRank}} & \footnotesize 0.0367 & \footnotesize 0.5451 & \footnotesize 0.4883 & \footnotesize 0.8884 & \footnotesize 0.1075 & \footnotesize 0.7549 & \footnotesize 0.0509 & \footnotesize 0.4446 & \footnotesize 0.0600 & \footnotesize 0.5900 & \footnotesize 0.1792 & \footnotesize 0.7820 \\
{\footnotesize \textbf{T5Sentinel}} & \footnotesize 0.0400 & \footnotesize 0.4621 & \footnotesize 0.0575 & \footnotesize 0.5519 & \footnotesize 0.0575 & \footnotesize 0.4580 & \footnotesize 0.0231 & \footnotesize 0.2582 & \footnotesize 0.0025 & \footnotesize 0.3473 & \footnotesize 0.0767 & \footnotesize 0.5286 \\
{\footnotesize \textbf{Binoculars}} & \footnotesize 0.3317 & \footnotesize 0.8930 & \footnotesize 0.7450 & \footnotesize 0.9364 & \footnotesize 0.4783 & \footnotesize 0.7969 & \footnotesize 0.1944 & \footnotesize 0.6989 & \footnotesize 0.1492 & \footnotesize 0.6754 & \footnotesize 0.5233 & \footnotesize 0.8807 \\
\bottomrule
\end{tabular}
}
\caption{Detector performance (AUROC and TPR@0.01) across tasks.}
\label{tab:detectorperf_other}
\end{table*}

\begin{table*}[!htbp]
\centering
\begin{tabular}{lcccccccc}

\toprule
  & \multicolumn{2}{c}{\footnotesize \textbf{QA EN}} & \multicolumn{2}{c}{\footnotesize \textbf{QA ES}} & \multicolumn{2}{c}{\footnotesize \textbf{QA FR}} & \multicolumn{2}{c}{\footnotesize \textbf{QA ZH}} \\
  & \footnotesize TPR & \footnotesize AUC & \footnotesize TPR & \footnotesize AUC & \footnotesize TPR & \footnotesize AUC & \footnotesize TPR & \footnotesize AUC \\
\midrule
{\footnotesize \textbf{Radar}} & \footnotesize 0.1225 & \footnotesize 0.7542 & \footnotesize 0.0100 & \footnotesize 0.4832 & \footnotesize 0.0525 & \footnotesize 0.5008 & \footnotesize 0.0400 & \footnotesize 0.3730 \\
{\footnotesize \textbf{Fast-DetectGPT}} & \footnotesize 0.5450 & \footnotesize 0.9063 & \footnotesize 0.4233 & \footnotesize 0.8522 & \footnotesize 0.6483 & \footnotesize 0.9390 & \footnotesize 0.5808 & \footnotesize 0.8864 \\
{\footnotesize \textbf{Wild}} & \footnotesize 0.1375 & \footnotesize 0.7437 & \footnotesize 0.0308 & \footnotesize 0.5019 & \footnotesize 0.0183 & \footnotesize 0.4660 & \footnotesize 0.0275 & \footnotesize 0.6180 \\
{\footnotesize \textbf{PHD}} & \footnotesize 0.0600 & \footnotesize 0.4694 & \footnotesize 0.0792 & \footnotesize 0.6228 & \footnotesize 0.0158 & \footnotesize 0.6371 & \footnotesize 0.0733 & \footnotesize 0.7138 \\
{\footnotesize \textbf{LogRank}} & \footnotesize 0.0725 & \footnotesize 0.7517 & \footnotesize 0.0725 & \footnotesize 0.6427 & \footnotesize 0.0200 & \footnotesize 0.7741 & \footnotesize 0.1308 & \footnotesize 0.8323 \\
{\footnotesize \textbf{T5Sentinel}} & \footnotesize 0.0558 & \footnotesize 0.6007 & \footnotesize 0.0075 & \footnotesize 0.4353 & \footnotesize 0.0025 & \footnotesize 0.4131 & \footnotesize 0.0175 & \footnotesize 0.6688 \\
{\footnotesize \textbf{Binoculars}} & \footnotesize 0.6950 & \footnotesize 0.9271 & \footnotesize 0.5292 & \footnotesize 0.8295 & \footnotesize 0.7250 & \footnotesize 0.9326 & \footnotesize 0.4908 & \footnotesize 0.8785 \\
\bottomrule
\end{tabular}
    \caption{Detector performance (AUROC and TPR@0.01) across multilingual QA tasks.}
    \label{tab:detectorperf_qa}
    \end{table*}

\begin{table*}[htbp]
\centering
 \begin{tabular}{lcccccccc}
\toprule
  & \multicolumn{2}{c}{\footnotesize \textbf{Summ EN}} & \multicolumn{2}{c}{\footnotesize \textbf{Summ ES}} & \multicolumn{2}{c}{\footnotesize \textbf{Summ FR}} & \multicolumn{2}{c}{\footnotesize \textbf{Summ ZH}} \\
  & \footnotesize TPR & \footnotesize AUC & \footnotesize TPR & \footnotesize AUC & \footnotesize TPR & \footnotesize AUC & \footnotesize TPR & \footnotesize AUC \\
\midrule
{\footnotesize \textbf{Radar}} & \footnotesize 0.0825 & \footnotesize 0.5721 & \footnotesize 0.0125 & \footnotesize 0.3839 & \footnotesize 0.0208 & \footnotesize 0.3966 & \footnotesize 0.2433 & \footnotesize 0.6940 \\
{\footnotesize \textbf{Fast-DetectGPT}} & \footnotesize 0.2175 & \footnotesize 0.8185 & \footnotesize 0.1408 & \footnotesize 0.6446 & \footnotesize 0.1058 & \footnotesize 0.6169 & \footnotesize 0.1950 & \footnotesize 0.6729 \\
{\footnotesize \textbf{Wild}} & \footnotesize 0.1250 & \footnotesize 0.7436 & \footnotesize 0.2683 & \footnotesize 0.6027 & \footnotesize 0.2350 & \footnotesize 0.5578 & \footnotesize 0.4392 & \footnotesize 0.8499 \\
{\footnotesize \textbf{PHD}} & \footnotesize 0.1292 & \footnotesize 0.5907 & \footnotesize 0.1417 & \footnotesize 0.6107 & \footnotesize 0.1617 & \footnotesize 0.5875 & \footnotesize 0.1933 & \footnotesize 0.6631 \\
{\footnotesize \textbf{LogRank}} & \footnotesize 0.7517 & \footnotesize 0.9705 & \footnotesize 0.4425 & \footnotesize 0.8902 & \footnotesize 0.4217 & \footnotesize 0.8754 & \footnotesize 0.4533 & \footnotesize 0.8317 \\
{\footnotesize \textbf{T5Sentinel}} & \footnotesize 0.1333 & \footnotesize 0.6275 & \footnotesize 0.0042 & \footnotesize 0.2728 & \footnotesize 0.0208 & \footnotesize 0.3219 & \footnotesize 0.0183 & \footnotesize 0.3828 \\
{\footnotesize \textbf{Binoculars}} & \footnotesize 0.3792 & \footnotesize 0.7916 & \footnotesize 0.2225 & \footnotesize 0.6866 & \footnotesize 0.1942 & \footnotesize 0.6482 & \footnotesize 0.1333 & \footnotesize 0.6935 \\
\bottomrule
\end{tabular}
    \caption{Detector performance (AUROC and TPR@0.01) across multilingual summarization tasks.}
    \label{tab:detectorperf_summ}
    \end{table*}

    \begin{table*}[htbp]
    \centering
\begin{tabular}{lcccccccc}
\toprule
  & \multicolumn{2}{c}{\footnotesize \textbf{Dialogue EN}} & \multicolumn{2}{c}{\footnotesize \textbf{Dialogue ES}} & \multicolumn{2}{c}{\footnotesize \textbf{Dialogue FR}} & \multicolumn{2}{c}{\footnotesize \textbf{Dialogue ZH}} \\
  & \footnotesize TPR & \footnotesize AUC & \footnotesize TPR & \footnotesize AUC & \footnotesize TPR & \footnotesize AUC & \footnotesize TPR & \footnotesize AUC \\
\midrule
{\footnotesize \textbf{Radar}} & \footnotesize 0.7175 & \footnotesize 0.9344 & \footnotesize 0.0583 & \footnotesize 0.6827 & \footnotesize 0.0417 & \footnotesize 0.5549 & \footnotesize 0.1317 & \footnotesize 0.6120 \\
{\footnotesize \textbf{Fast-DetectGPT}} & \footnotesize 0.7475 & \footnotesize 0.9508 & \footnotesize 0.4300 & \footnotesize 0.8847 & \footnotesize 0.6158 & \footnotesize 0.9098 & \footnotesize 0.4217 & \footnotesize 0.9011 \\
{\footnotesize \textbf{Wild}} & \footnotesize 0.0292 & \footnotesize 0.8720 & \footnotesize 0.0775 & \footnotesize 0.6266 & \footnotesize 0.1000 & \footnotesize 0.6035 & \footnotesize 0.0617 & \footnotesize 0.3290 \\
{\footnotesize \textbf{PHD}} & \footnotesize 0.0425 & \footnotesize 0.6224 & \footnotesize 0.1892 & \footnotesize 0.7168 & \footnotesize 0.1842 & \footnotesize 0.7401 & \footnotesize 0.3835 & \footnotesize 0.7962 \\
{\footnotesize \textbf{LogRank}} & \footnotesize 0.8583 & \footnotesize 0.9889 & \footnotesize 0.4250 & \footnotesize 0.8824 & \footnotesize 0.4283 & \footnotesize 0.8987 & \footnotesize 0.5208 & \footnotesize 0.8870 \\
{\footnotesize \textbf{T5Sentinel}} & \footnotesize 0.1708 & \footnotesize 0.6628 & \footnotesize 0.0467 & \footnotesize 0.6010 & \footnotesize 0.0508 & \footnotesize 0.4639 & \footnotesize 0.0217 & \footnotesize 0.3446 \\
{\footnotesize \textbf{Binoculars}} & \footnotesize 0.7767 & \footnotesize 0.9489 & \footnotesize 0.7700 & \footnotesize 0.9407 & \footnotesize 0.7125 & \footnotesize 0.9393 & \footnotesize 0.6467 & \footnotesize 0.9475 \\
\bottomrule
\end{tabular}
    \caption{Detector performance (AUROC and TPR@0.01) across multilingual dialogue tasks.}
    \label{tab:detectorperf_dialogue}
    \end{table*}


